\documentclass[lettersize,journal]{IEEEtran}
\usepackage{amsmath,amsfonts}
\usepackage{algorithmic}
\usepackage{algorithm}
\usepackage{array}
\usepackage[caption=false,font=footnotesize]{subfig}
\usepackage{textcomp}
\usepackage{stfloats}
\usepackage{url}
\usepackage{verbatim}
\usepackage{graphicx}
\usepackage{cite}
\begin{document}

\title{A Pedestrian–Vehicle Interaction Benchmark and Annotation Framework for Unstructured Scenes via Uncalibrated Cameras}

\author{Haoyang Peng, Qian Hu, Songan Zhang$^{*}$, Ming Yang~\IEEEmembership{Member,~IEEE,}
\thanks{This work was supported in part by the National Natural Science Foundation of China under Grant 52402504 and Grant 62473250. \textit{Corresponding author: Songan Zhang, e-mail: songanz@sjtu.edu.cn.}}
\thanks{Haoyang Peng and Ming Yang are with the School of Automation and Intelligent Sensing, Shanghai Jiao Tong University, Shanghai 200240, China.}
\thanks{Qian Hu and Songan Zhang are with the Global Institute of Future Technology, Shanghai Jiao Tong University, Shanghai 200240, China (e-mail: songanz@sjtu.edu.cn).}
}

\markboth{Journal of \LaTeX\ Class Files,~Vol.~14, No.~8, August~2021}%
{Shell \MakeLowercase{\textit{et al.}}: A Sample Article Using IEEEtran.cls for IEEE Journals}


\maketitle

\begin{abstract}
Predicting the interaction between pedestrian and vehicle is essential for autonomous driving safety in unstructured and semi-structured scenarios; however, this task is severely hindered by the scarcity of public datasets that feature dense pedestrian-vehicle interactions. Most current studies rely on structured road data, leaving the complex, heterogeneous interactions found in unstructured environments insufficiently represented and researched. In this paper, we propose a dataset annotation framework based on video data from uncalibrated surveillance cameras and present PINNS (Pedestrian–vehicle Interaction dataset from uNcalibrated cameras in uNstructured Scenes). The dataset covers multiple countries and regions, includes diverse typical traffic scenarios, and considers variations in seasons, lighting conditions, and weather. It focuses on complex scenes with dense pedestrian–vehicle interactions and is designed to be easily extensible. The dataset is constructed and annotated according to the standard issued by the Chinese Association of Automation, providing both trajectory data and corresponding scene-level information. Furthermore, this paper analyzes current challenges and research directions in heterogeneous agent trajectory prediction, shows the necessity and usefulness of the proposed dataset. We hope our framework and dataset will facilitate research on trajectory prediction and autonomous driving in complex mixed traffic scenarios. PINNS is publicly available at \url{https://github.com/Songan-Lab}.

\end{abstract}


\section{Introduction}
\IEEEPARstart{T}{raffic} accidents involving pedestrians constitute a severe global safety challenge, with pedestrians alone accounting for 21\%\cite{accidentamount} of all traffic fatalities. Unlike vehicle-vehicle collisions, these accidents disproportionately result in severe injuries or death due to the lack of physical protection for pedestrians\cite{injuryrate}. Consequently, ensuring the safety of these participants has become a critical prerequisite and major challenge for the large-scale deployment of autonomous driving systems. Understanding and predicting road user interactions is fundamental to accident prevention in autonomous driving. While data-driven deep learning has become the dominant paradigm for this task, the efficacy of these models relies heavily on exposure to diverse real-world scenarios. Consequently, a comprehensive dataset capturing rich pedestrian-vehicle interactions is indispensable. Such data is critical not only for training robust models that generalize to mixed traffic environments but also for benchmarking their safety and performance.

Compared with relatively well-studied vehicle–vehicle or pedestrian–pedestrian interactions, interactions between pedestrians and vehicles show fundamentally different and more complex characteristics. This complexity arises from intrinsic differences between the two agents in terms of kinematic constraints, feasible motion regions, behavioral logic, and decision-making principles.Vehicles generally follow explicit dynamic constraints and traffic rules, resulting in relatively continuous and predictable motion patterns. In contrast, pedestrians exhibit higher flexibility and uncertainty, as their movements are influenced by environmental perception, subjective risk assessment, and social interactions. In real traffic scenarios, pedestrians often show clear yielding behavior toward vehicles, and their perception of vehicle speed, distance, and intent differs fundamentally from the interaction and game mechanisms between vehicles\cite{pedactionmode}. However, in certain regions or scenarios, social factors such as traffic regulations or contextual constraints may significantly influence interaction behaviors, leading to opposite phenomena. Unlike structured road environments with strict separation between pedestrians and vehicles, semi-structured and unstructured shared spaces contain the majority of pedestrian–vehicle interactions. The trajectory and behavior characteristics of pedestrians and vehicles differ significantly across structured and unstructured environments, as well as among different types of unstructured scenarios\cite{trajfeature}. Modeling, prediction, and analysis of such complex and diverse interactions require trajectory datasets with sufficient diversity, coverage, and richness in unstructured road scenarios.

According to existing studies, publicly available trajectory prediction datasets primarily fall into two categories. The first category consists of general pedestrian trajectory prediction datasets, such as ETH and UCY\cite{eth,ucy}. These datasets are typically collected in pedestrian-only environments, such as campuses, shopping malls, or plazas, and lack realistic traffic scenarios. The second category includes traffic scene datasets designed for autonomous driving, such as nuScenes, Waymo, KITTI,  Argoverse\cite{nuscenes, waymo, kitti, argoverse}. These datasets contain large amounts of traffic data but primarily focus on vehicle behavior and are mostly collected in structured road or highway environments. Some datasets target unstructured or semi-structured scenarios, such as SDD, inD, DLP, INTERACTION, CITR and DUT\cite{sdd, ind, dlp, interaction, citrdut}. However, these datasets generally have limited coverage and scale, often focusing on specific locations or regions. Datasets such as INTERACTION provide trajectory data from semi-structured environments across different regions, but their primary focus remains vehicle–vehicle interactions. Overall, there is still a lack of an open dataset that focuses on pedestrian–vehicle interactions in unstructured road environments and provides sufficient diversity in terms of geographic distribution, scene types, and temporal coverage.

From a data collection perspective, advances in sensing technology have made it possible to construct large-scale, high-quality datasets. Most large traffic datasets\cite{argoverse, nuscenes, waymo, apollo, lyft, kitti} rely on vehicle-mounted sensors, including cameras, LiDAR, radar, and GPS, to collect multimodal and high-precision data. The dataset scale can be conveniently expanded by increasing vehicle mileage. Some datasets\cite{sdd,citrdut,ind,onsitevru} use unmanned aerial vehicles (UAVs) to monitor and record designated traffic scenes. UAVs can avoid occlusions and minimize the influence of sensing equipment on observed behaviors, while also enabling long-term fixed-point observation. However, most existing traffic datasets depend on professional equipment such as vehicle-mounted sensors or UAVs, which require precise calibration and deployment conditions. Although these methods provide high-accuracy data, they inherently limit geographic coverage, scene diversity, and long-term scalability. Vehicle-mounted data collection is constrained by vehicle accessibility and is more likely to capture structured rather than unstructured road environments. UAVs and calibrated roadside cameras require fixed deployment and are sensitive to weather and lighting conditions, which restricts diverse and continuous data collection. 

In contrast, data collected from non-specialized devices, such as widely deployed surveillance cameras, are easier to obtain and offer broad coverage of unstructured road environments. Although such data typically lack explicit calibration parameters and depth information, recent advances in monocular 3D estimation\cite{3destimation1, 3destimation2} and camera self-calibration\cite{autocali1, autocali2, autocali3} have made it increasingly feasible to recover useful geometric information directly from video data. Given the extensive availability of uncalibrated surveillance cameras, these approaches present significant potential for scalable data collection in unstructured environments.

In this paper, to solve the lack of trajectory prediction datasets for unstructured pedestrian–vehicle interaction scenarios and to promote research on trajectory prediction and autonomous driving in mixed traffic environments, we propose a dataset construction framework based on uncalibrated surveillance camera data. Using this framework, we build a pedestrian–vehicle interaction trajectory prediction dataset covering multiple countries and regions, diverse scene types, and different time periods. Our main contributions are summarized as follows:
\begin{enumerate}
    \item We propose a trajectory prediction dataset annotation framework and pipeline based on uncalibrated surveillance camera data, and build PINNS, an unstructured road pedestrian–vehicle interaction trajectory dataset following this framework.
    \item PINNS provides rich temporal and spatial diversity, covering multiple countries and regions, various types of unstructured road scenes, different seasons, different times of day, and diverse lighting and weather conditions, including sunny, rainy, and snowy scenarios.
    \item PINNS focuses on semi-structured and unstructured roads with dense pedestrian–vehicle interactions. We filter out scenes without meaningful interactions and retain complex interaction scenarios for further processing and annotation.
    \item We evaluate representative trajectory prediction methods on the proposed dataset and reveal the limitations of existing approaches in complex mixed traffic scenarios.
\end{enumerate}

The remainder of this paper is organized as follows. Section \ref{sec:2} reviews existing trajectory prediction datasets, including general pedestrian datasets and traffic datasets designed for autonomous driving, and summarizes representative trajectory prediction and behavior modeling approaches. Section \ref{sec:3} introduces the proposed unstructured road pedestrian-vehicle interaction trajectory dataset, including its construction framework and key characteristics. Section \ref{sec:4} presents quantitative evaluations of baseline trajectory prediction methods on the proposed dataset. Finally, Section \ref{sec:5} concludes the paper and discusses future research directions

\section{Related Work}
\label{sec:2}
In this section, we review existing datasets for trajectory prediction in autonomous driving, with a focus on pedestrian–vehicle interaction scenarios. We then summarize representative approaches for trajectory prediction and behavior modeling, emphasizing recent methods that model interactions between heterogeneous agents.

\subsection{Pedestrian–Vehicle Interaction Datasets for Autonomous Driving}
With the rapid development of autonomous driving, large-scale traffic datasets have become the dominant data source. Datasets such as nuScenes\cite{nuscenes}, Argoverse\cite{argoverse}, and the Waymo Open Dataset\cite{waymo} have achieved significant progress in terms of data scale, sensor and annotation diversity using onboard multi-sensor platforms. Such sensor configurations enable accurate and multimodal spatiotemporal perception in complex urban environments, providing strong support for vehicle trajectory prediction and related tasks such as object detection and tracking. However, because these datasets are constructed from an ego-vehicle perspective, the collected scenes are mostly limited to structured road environments. Pedestrian–vehicle interaction samples account for only a small portion of the data and usually occur under well-defined traffic rules, which limited suitability for studying complex interactions in unstructured or shared spaces.

In contrast to large-scale traffic datasets collected in structured scenes using onboard sensors, several recent datasets have shifted their focus toward unstructured environments observed by fixed or external sensing devices. Unmanned aerial vehicles (UAVs) have been widely adopted to overcome the limitations of onboard sensing and to enable continuous observation of specific areas for these datasets. The Stanford Drone Dataset (SDD)\cite{sdd} uses UAVs to record overhead videos of campus environments. The INTERACTION dataset\cite{interaction} further expands geographic coverage. Although these datasets contain some mixed pedestrian–vehicle scenes, their primary focus remains on either pedestrian behavior or vehicle interactions, with limited scene diversity. The CITR and DUT datasets\cite{citrdut} collect pedestrian-vehicle interaction trajectory using UAVs in campus and controlled scenarios. The Dragon Lake Parking dataset\cite{dlp} and OnSiteVRU\cite{onsitevru} dataset focus on specific unstructured scenarios. While these datasets partially address the lack of pedestrian–vehicle interaction data in low-speed, unstructured environments, they are still limited to specific regions, scenario types or data scale, which remain insufficient to support large-scale generalization studies. Moreover, data collection methods relying on UAVs or specialized sensing equipment are more sensitive to weather and deployment conditions, which further limits scalability and diversity.

Overall, unstructured environments and pedestrian–vehicle interactions are often not the primary design targets, resulting in limited data availability. In addition, most existing datasets depend on expensive and calibrated professional sensing equipment, and their deployment cost and difficulty restrict the expansion of geographic and scenario coverage. Therefore, our work aims to construct a more comprehensive dataset for pedestrian–vehicle interactions in unstructured environments and to propose a dataset construction framework that does not rely on specialized sensing equipment, enabling more convenient data collection and long-term scalability.

\begin{table*}[t]
\caption{Comparison of representative trajectory prediction datasets with PINNS.}
\label{tab:datasets}
\centering
\small
\setlength{\tabcolsep}{4pt}
\renewcommand{\arraystretch}{1.2}
\begin{tabular}{p{2cm} c p{1.6cm} p{2.2cm} p{2.0cm} p{2cm} c p{1.5cm} p{1.5cm}}
\hline
Dataset &
Year &
Country/ Region &
Main Scenario &
Weather / Season / Night &
Acquisition Method &
Unstructured &
Observed Agents &
Interaction Type \\
\hline
ETH\cite{eth} &
2008 &
Switzerland &
Urban sidewalks / plazas &
- &
Fixed camera &
Yes &
ped. &
ped.--ped. \\

UCY\cite{ucy} &
2010 &
Cyprus &
Campus plazas / walkways &
- &
Fixed camera &
Yes &
ped. &
ped.--ped. \\

SDD\cite{sdd} &
2016 &
USA &
Campus, intersections &
- &
UAVs &
Yes &
ped. (dom.) &
ped.--ped. \\

HBS\cite{hbs} &
2017 &
Germany &
Shared space &
- &
Fixed camera &
Yes &
ped. veh. &
ped.--veh. \\

CITR\cite{citrdut} &
2019 &
USA &
Campus roads &
- &
UAVs &
Yes &
ped. veh. &
ped.--veh. \\

DUT\cite{citrdut} &
2019 &
China &
Campus &
- &
UAVs &
Yes &
ped. veh. &
ped.--veh. \\

PIE\cite{pie}&
2018 &
Multiple &
Urban roads &
Multi-weather &
On-board camera &
No &
ped. &
ped. to veh. \\

nuScenes\cite{nuscenes} &
2019 &
USA, Singapore &
Urban roads &
Multi-weather / night &
On-board multi-sensors &
No &
veh. (ego) &
- \\

INTERACTION\cite{interaction} &
2021 &
Multiple &
Intersections, roundabouts, highway merging &
Multi-weather &
UAVs / camera &
No &
veh. &
veh.--veh. \\

DLP\cite{dlp} &
2022 &
China &
Parking lot &
- &
UAVs &
Yes &
veh. &
veh.--veh. \\

\textbf{PINNS} &
- &
\textbf{Multiple} &
\textbf{Multiple unstructured scenarios} &
\textbf{Multi-weather / multi-seasons / night} &
\textbf{Uncalibrated cameras} &
\textbf{Yes} &
\textbf{ped. veh.} &
\textbf{ped.--veh.} \\
\hline
\end{tabular}
\par
\smallskip
\begin{minipage}{0.967\textwidth} 
    \footnotesize 
    Abbreviations and Legend: ped.\ = pedestrian, veh.\ = "A--B" denotes symmetric interactions; "A to B" denotes asymmetric behavioral response. "-" denotes unavailable information.
\end{minipage}
\end{table*}

\begin{figure*}[t]
    \centering
    \includegraphics[width=.87\linewidth]{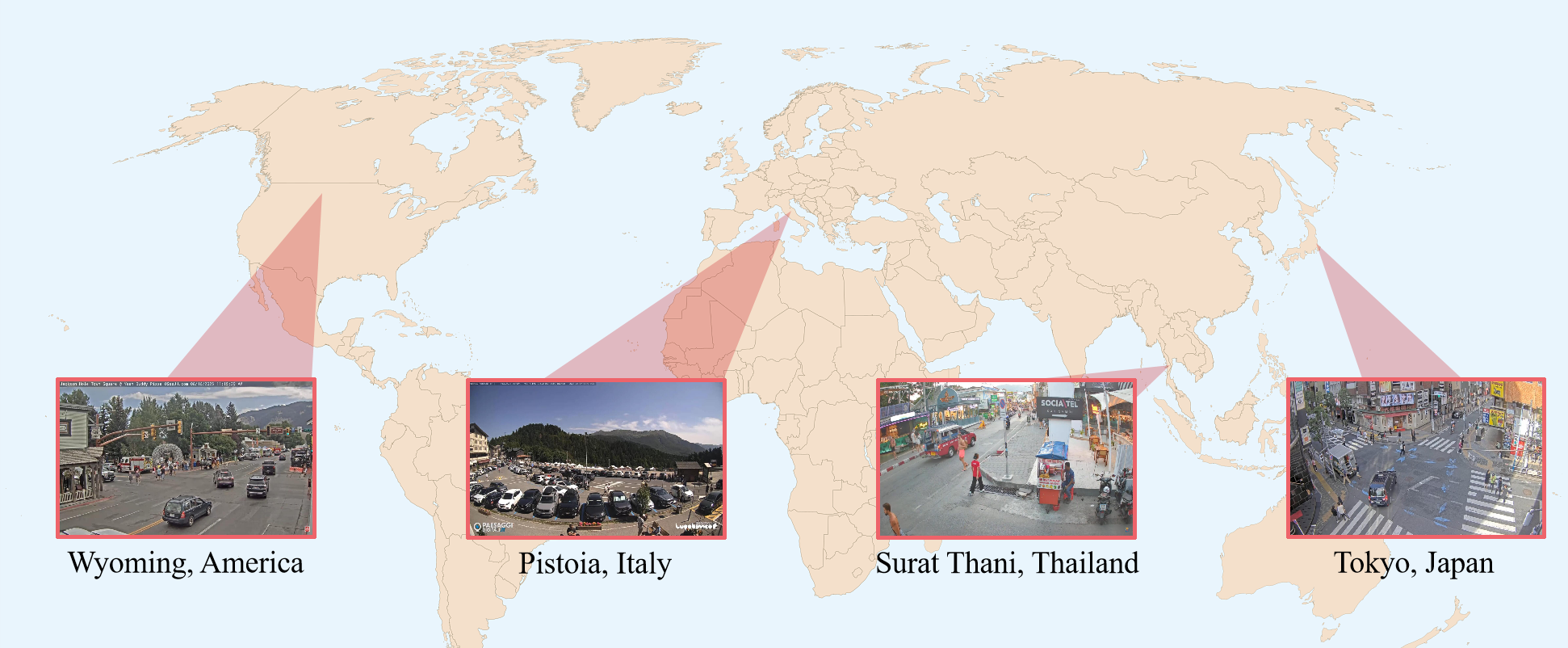}
    \caption{Geographical distribution of dataset scenes.}
    \label{fig:2}
\end{figure*}

\begin{figure}[t]
    \centering
    \includegraphics[width=\linewidth]{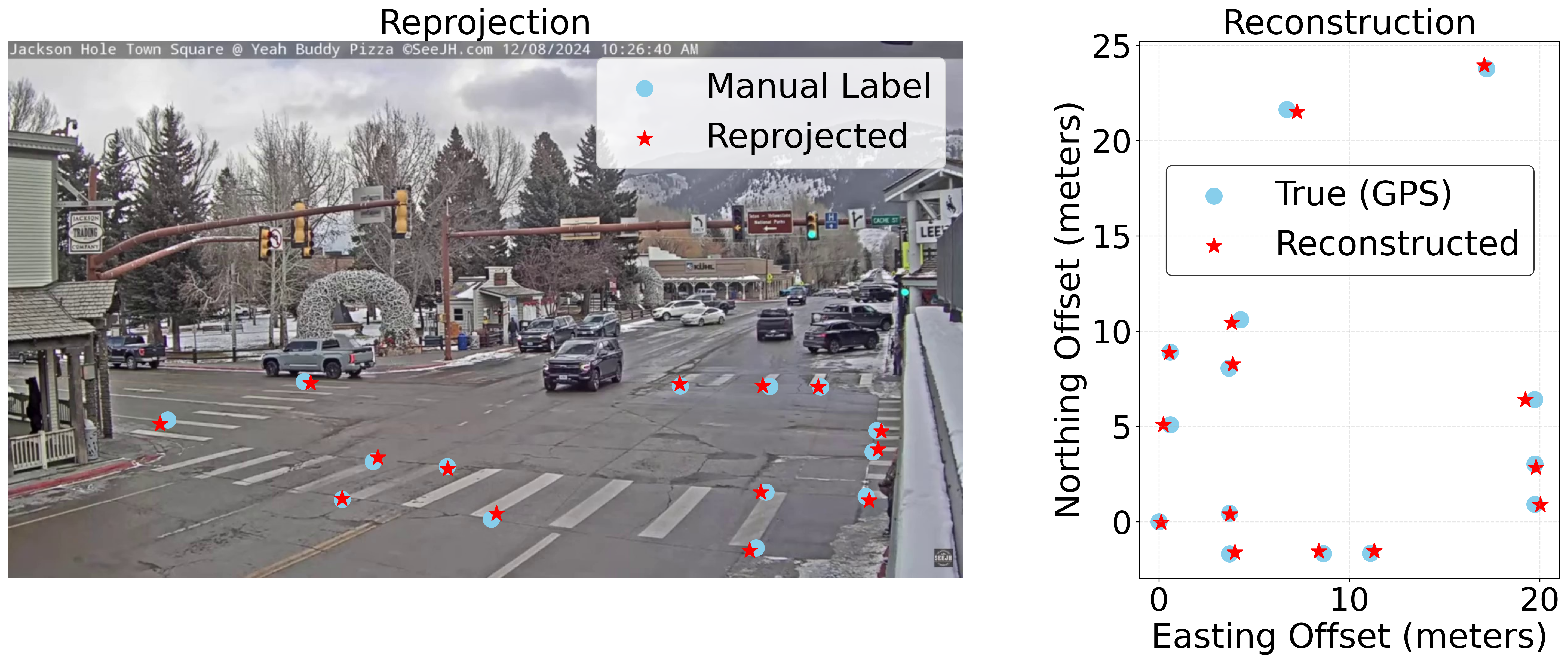}
    \caption{Reprojection and reconstruction visualization of the homography calibration method in a sample scene. Left: Reprojection of world points onto the image plane; Right: Reconstruction of image points into the metric world coordinate system. Blue dots represent the original reference points, while red dots indicate the corresponding reprojected or reconstructed points.}
    \label{fig:5}
\end{figure}

\subsection{Trajectory Prediction and Behavior Modeling Methods}
In trajectory prediction, methods focusing on a single type of agent have been extensively studied. Early trajectory prediction approaches mainly relied on physics-based and kinematic models, including social force model, velocity obstacle methods, crowd models, and collision avoidance models\cite{socialforce,vo,rvo,crowdmodel,occupancy}. These methods are interpretable and perform reasonably well in simple environments, but their modeling capacity is limited by handcrafted models, making them difficult to generalize to complex traffic scenarios. With the increasing availability of data and advances in deep learning, learning-based methods have gradually surpassed traditional physics-based approaches. In particular, sequence modeling frameworks based on recurrent neural networks (RNNs) became dominant, with representative pedestrian prediction methods such as Social LSTM \cite{sociallstm} and Social GAN \cite{socialgan} introducing social pooling mechanisms to capture inter-agent interactions, followed by graph- and spatial-aware extensions such as SGCN \cite{sgcn} and SocialCircle \cite{socialcircle}. Overall, these single-agent-type methods have laid a solid foundation for trajectory prediction but remain limited in modeling heterogeneous interactions in complex traffic environments.

In complex traffic environments with mixed agent types, heterogeneous multi-agent interactions and environmental constraints have become key research focus. Trajectron++\cite{trajectron} models interactions among different agent types using a unified heterogeneous graph representation. TraPHic\cite{traphic} and TrafficPredict\cite{trafficpredict} introduce weighted interaction mechanisms to handle heterogeneous traffic participants. Most of these approaches use LSTM-based temporal encoders as the core prediction module, combined with graph-based or attention-based interaction modeling.

More recently, Transformer architectures and diffusion models have been introduced into trajectory prediction and have gradually become core modeling frameworks. Methods such as Scene Transformer\cite{scenetransformer}, QCNet\cite{qcnet}, SIMPL\cite{simpl}, and MultiPath++\cite{multipath} leverage Transformer-based architectures to jointly model multi-agent interactions and environmental context. ParkDiffusion\cite{dlp} adopts diffusion models to capture the inherent uncertainty and multimodality of future trajectories. While these methods achieve promising performance in specific scenarios and show strong capability in modeling vehicle–vehicle interactions, dataset limitations remain a key bottleneck for advancing heterogeneous pedestrian–vehicle interaction modeling. The performance of existing models is highly dependent on the datasets and scenarios they are trained and evaluated on, which are still dominated by vehicle-vehicle interactions and limited patterns of pedestrian–vehicle interaction.

\section{Dataset}
\label{sec:3}

\begin{figure*}[t]
    \centering
    \includegraphics[width=0.9\linewidth]{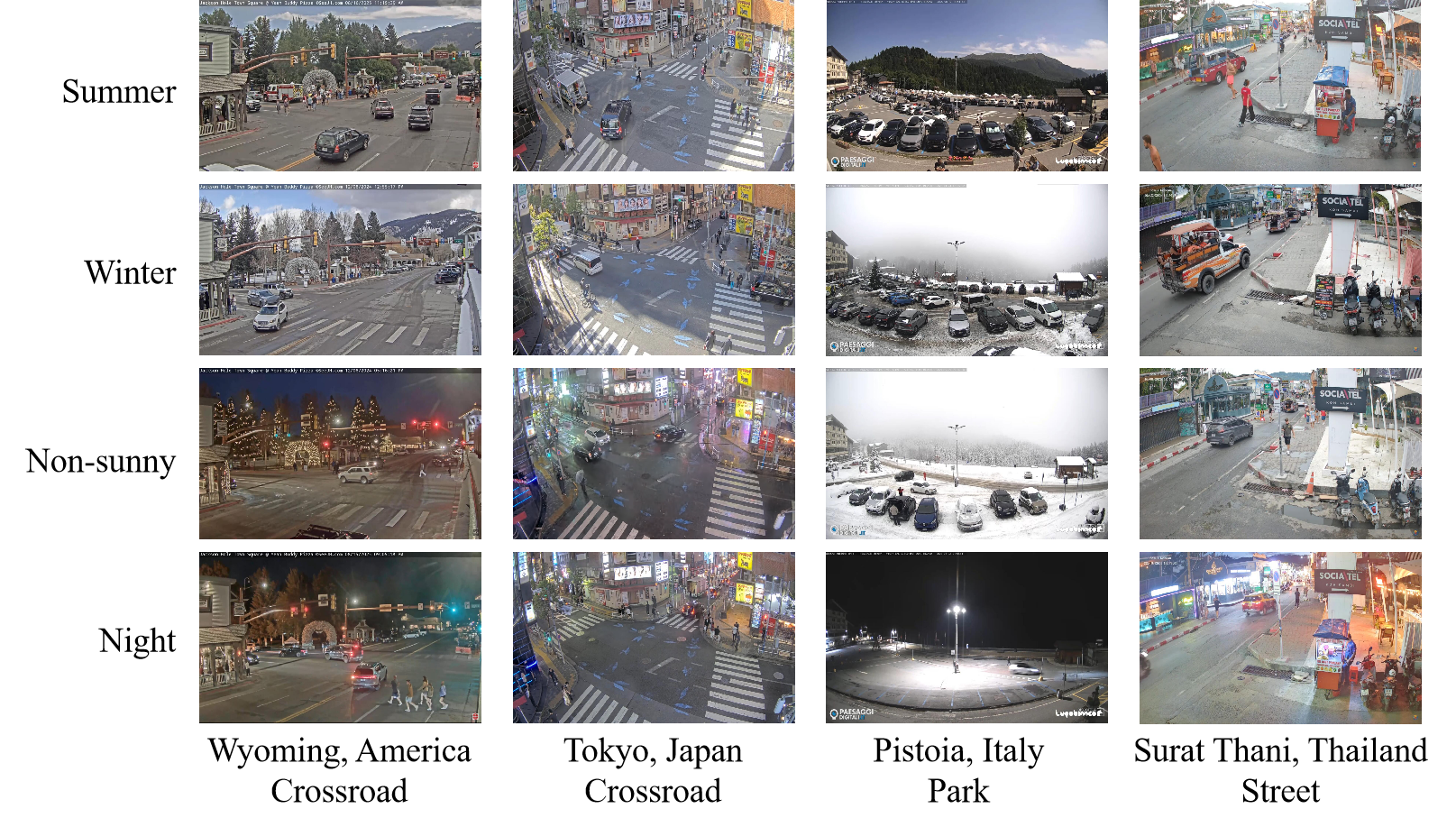}
    \caption{Interaction scene images under different environments. Vertical axis corresponds to different seasons, times of day, and weather conditions, while the horizontal axis indicates scene locations and types.}
    \label{fig:3}
\end{figure*}

In this section, we present the overall construction process and characteristics of the proposed pedestrian–vehicle interaction trajectory prediction dataset. The dataset is constructed in accordance with the trajectory prediction dataset standard issued by the Chinese Association of Automation \cite{TCAA2022002}. The comparison in Table \ref{tab:datasets} shows that PINNS offers clear advantages in terms of the richness of unstructured pedestrian–vehicle interaction scenes and environmental diversity. 

\subsection{Data Collection and Scene Distribution}

Unlike most existing traffic datasets that are collected using onboard sensors such as cameras and LiDAR or specialized platforms such as UAVs, PINNS is constructed from uncalibrated traffic surveillance cameras installed at fixed viewpoints above the scenes. This design choice is motivated by the observation that low-speed, complex, and unstructured pedestrian–vehicle interactions tend to occur in specific localized areas, such as shared spaces, intersections, campus roads, or internal roads in commercial areas, rather than along continuous and standardized structured road networks. For these scattered and diverse unstructured scenes, compared with onboard sensors, fixed surveillance cameras can continuously and stably observe the same location over long periods without being constrained by vehicle trajectories. Moreover, these unstructured scene data can be easily obtained from publicly available online surveillance cameras. Compared with datasets that rely on specialized equipment for data collection, surveillance-based data acquisition has lower cost and better scalability.

The data sources of PINNS consist of open-source online surveillance videos \cite{italy_webcam, japan_webcam, america_webcam, thailand_webcam}. The spatial distribution of the scenes is shown in Fig. \ref{fig:2}. The dataset covers multiple cities worldwide and different types of road environments. In total, the dataset contains 4 representative scenes from different regions, including intersections, streets, and parking areas that function as unstructured shared spaces. These scenes exhibit significant diversity in geographic location and road layout.

All videos are recorded at a resolution of $1536 \times 864$ with a frame rate of 30 fps. The raw data consist of more than 6000 minutes videos in total, spanning different seasons within a year. The dataset includes various weather conditions, such as sunny, rainy, and snowy days, as well as different illumination conditions, including daytime and nights. Fig. \ref{fig:3} shows the diversity of scenes and environmental conditions in the dataset.

\subsection{Scene Calibration and Data Filtering}

\begin{figure*}[t]
    \centering
    \includegraphics[width=.8\linewidth]{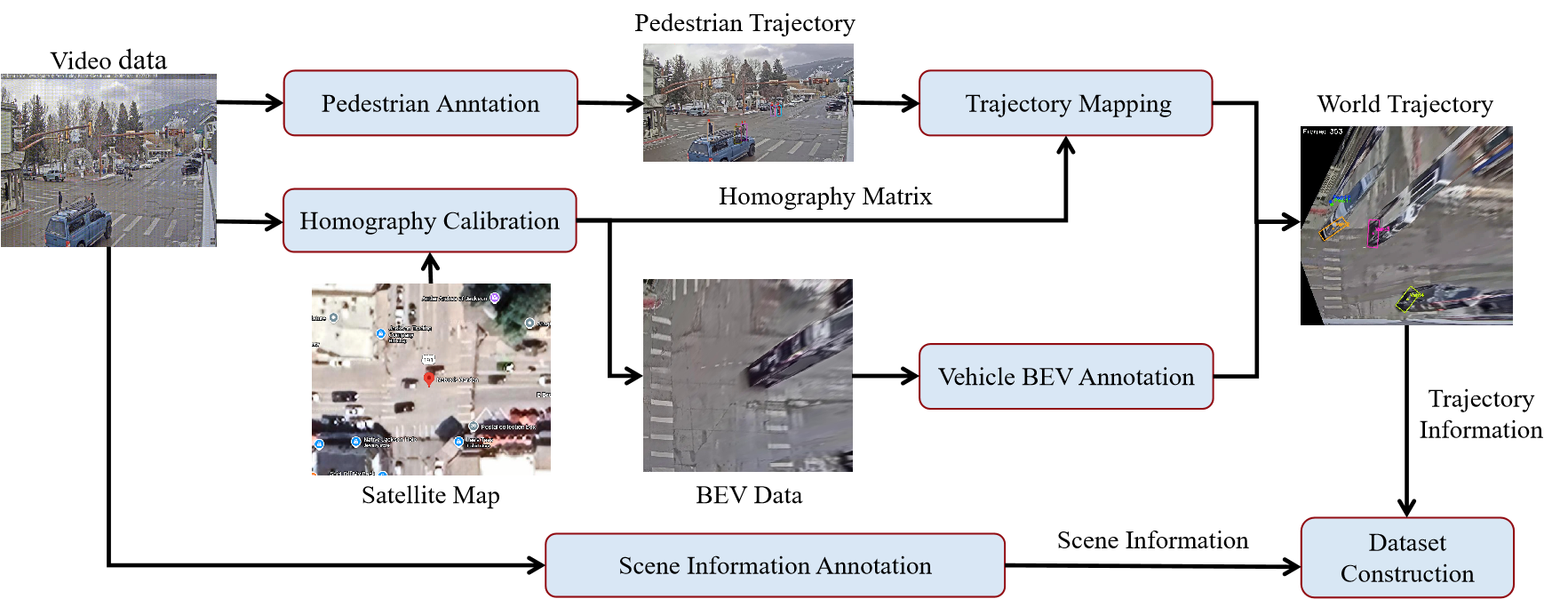}
    \caption{Overview of the dataset construction pipeline.}
    \label{fig:1}
\end{figure*}

Since most collected videos lack depth information and accurate camera intrinsic and extrinsic are unavailable, recovering trajectories in the real-world coordinate system from monocular image sequences is a major challenge. To solve this issue, we propose a scene calibration and trajectory transformation approach based on the bird’s-eye view (BEV).

Specifically, as shown in Fig. \ref{fig:1}, we first localize each scene using high-resolution satellite maps to determine a local reference plane in the world coordinate system. Then, by selecting corresponding points between the image plane and the map plane, a homography matrix is computed to project object positions from the original image plane to the BEV. This process yields approximate 2D planar trajectories in the real world based on the image plane trajectory. Since pedestrian–vehicle interaction scenes can generally be approximated as locally planar, the resulting BEV trajectories can well preserve scale and geometric relationships and reflect real motion characteristics.

Compared with traditional calibration methods that rely on precise camera intrinsic and extrinsic, this approach avoids the complexity of full camera calibration for unknown cameras and is more tolerant to calibration and annotation errors. In practice, an average of 10 to 15 reference points are selected per scene to solve the overdetermined homography system. Quantitative evaluation indicates that at an original resolution of $1536 \times 864$, the average reprojection error is 8.24 pixels, with a corresponding mean reconstruction error of 0.28 meters. The visualization of reprojection and reconstruction for a representative scene is presented in Fig. \ref{fig:5}. Since the method maps coordinates directly onto a 2D manifold without requiring depth estimation, any minor projection offsets exhibit spatial consistency and remains controllable. Consequently, these deviations do not compromise the integrity of trajectory attributes or relative motion relationships, ensuring the reliability of subsequent motion analysis and trajectory prediction.

\begin{figure*}[t]
    \centering
    \subfloat[Location distribution]{
        \includegraphics[width=0.23\linewidth]{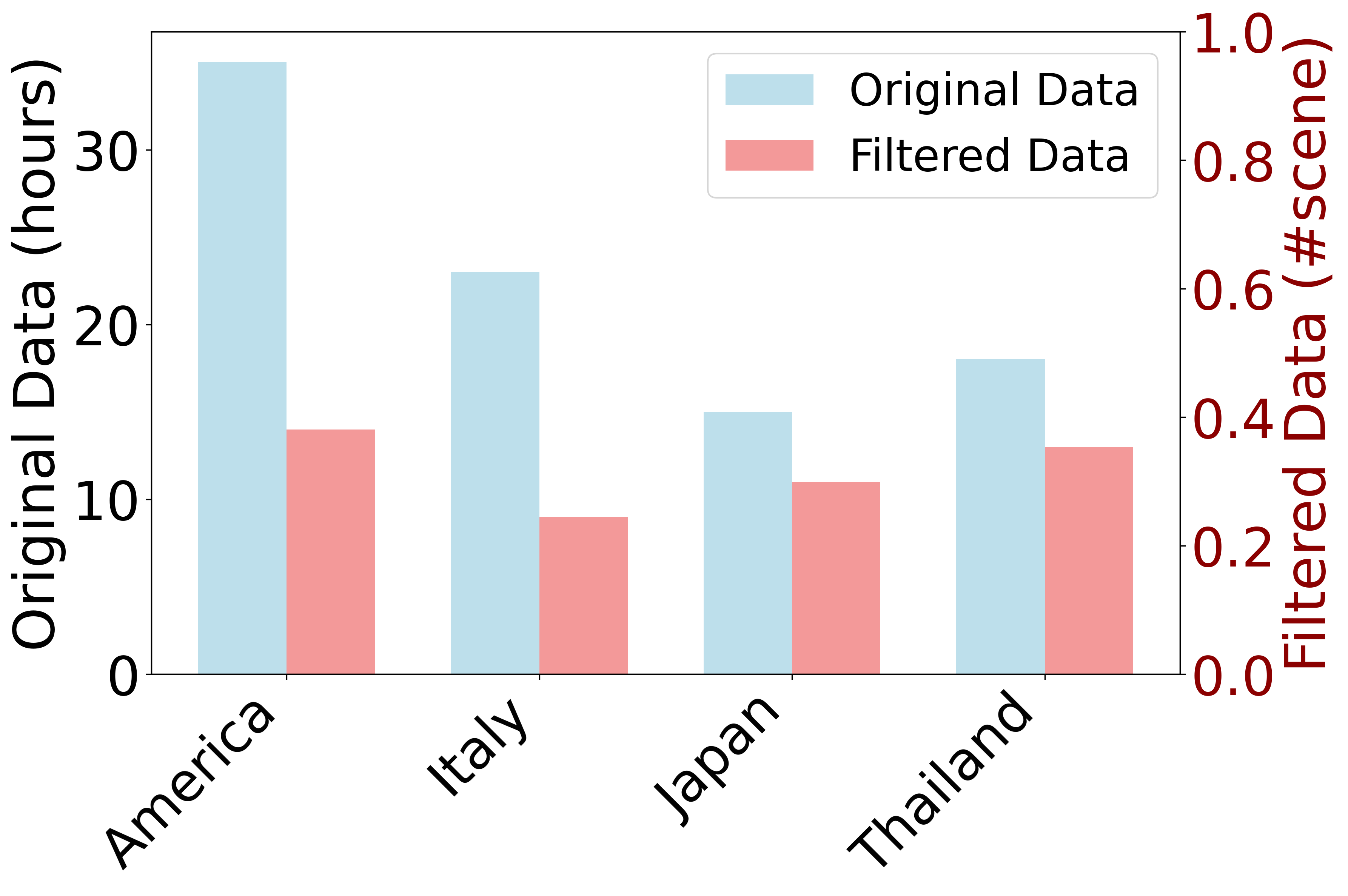}
    }
    \hfill
    \subfloat[Season distribution]{
        \includegraphics[width=0.23\linewidth]{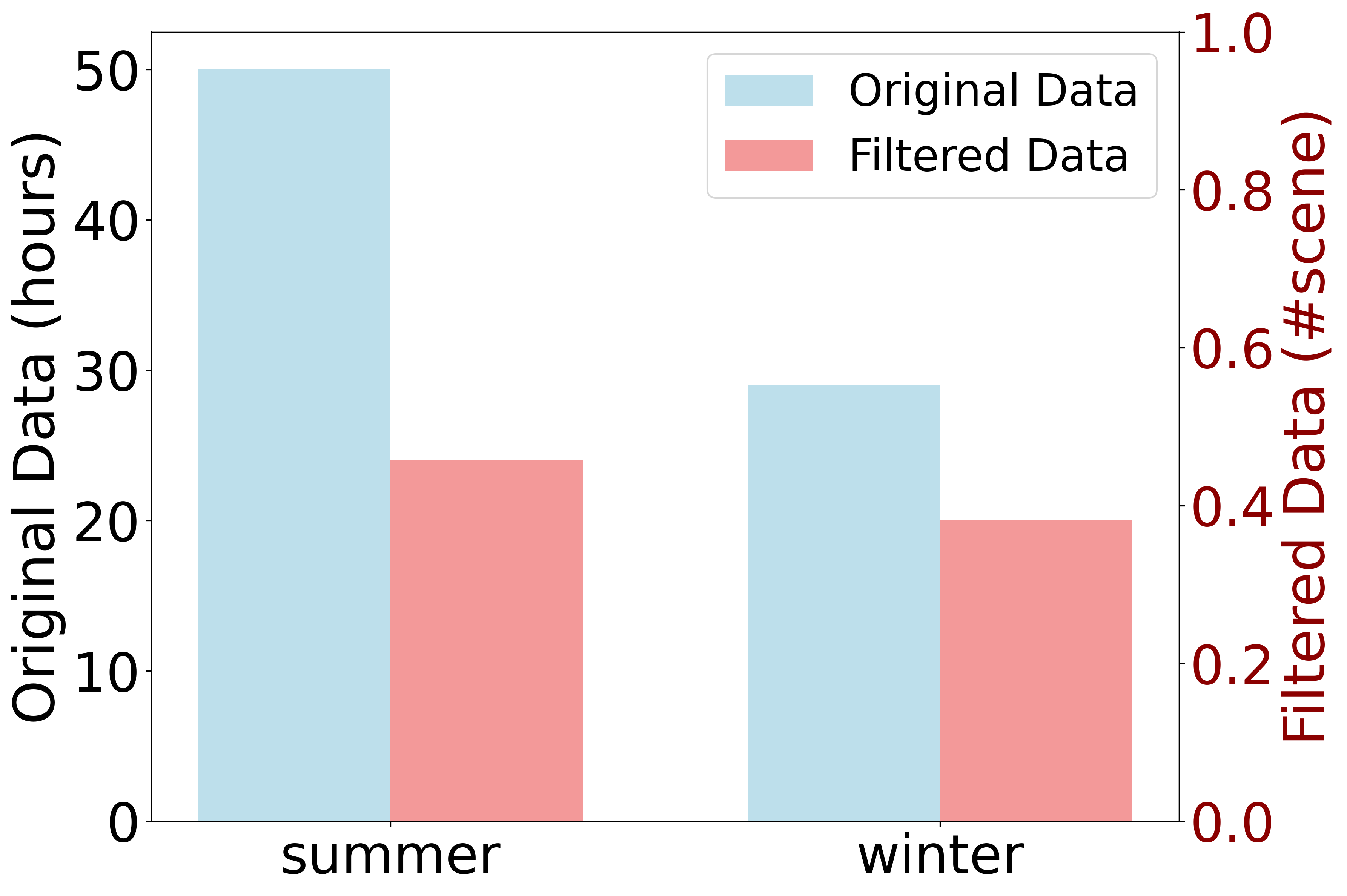}
    }
    \hfill
    \subfloat[Time-of-day distribution]{
        \includegraphics[width=0.23\linewidth]{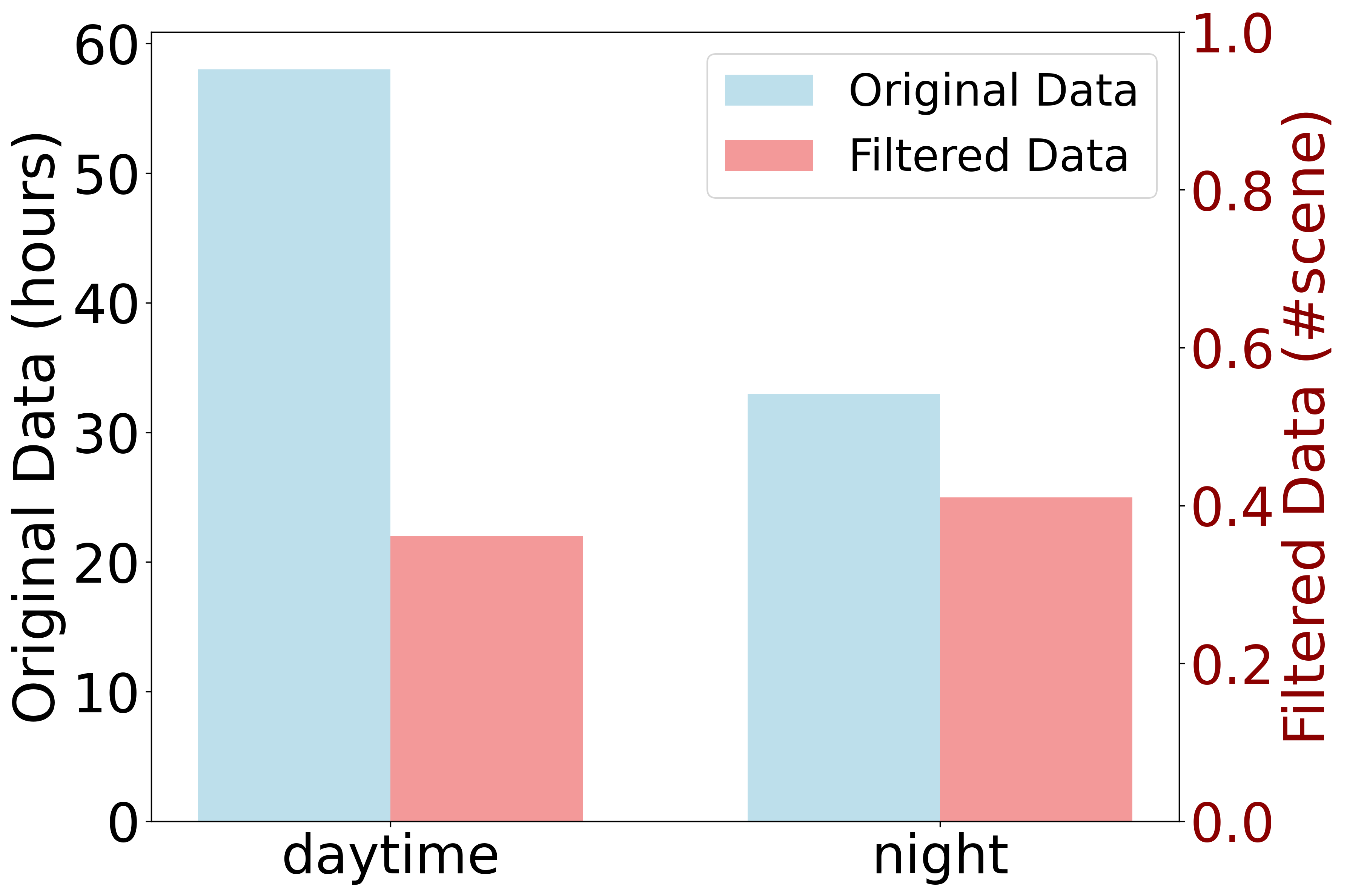}
    }
    \hfill
    \subfloat[Weather distribution]{
        \includegraphics[width=0.23\linewidth]{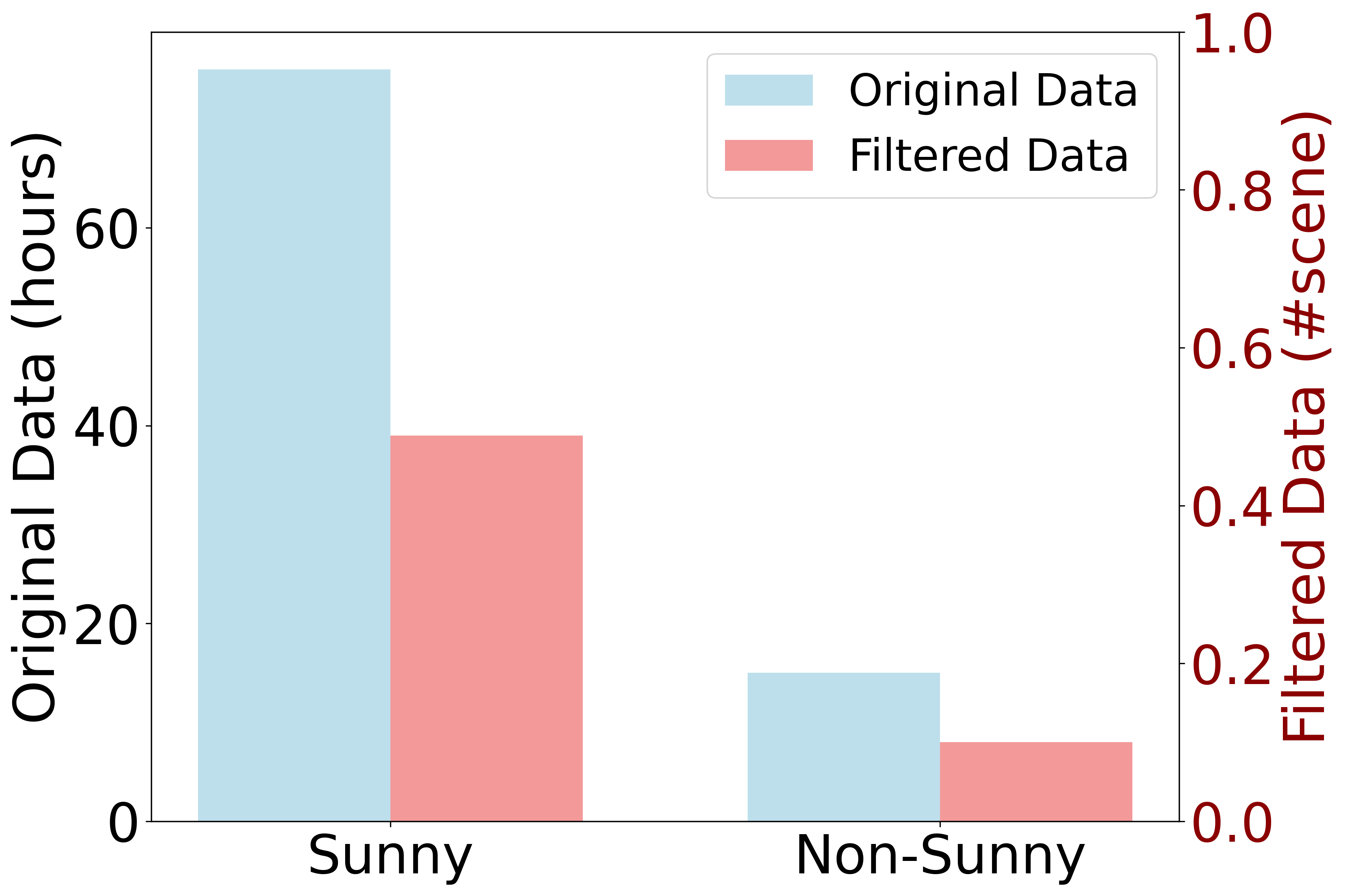}
    }
    \caption{Comparison of data distributions between the raw data and the filtered dataset. Raw scenes are counted by recording hours, while the filtered dataset is counted by the number of scenes.}
    \label{fig:4}
\end{figure*}

\begin{figure*}[t]
\centering
\subfloat[Location distribution]{
  \includegraphics[width=0.23\linewidth]{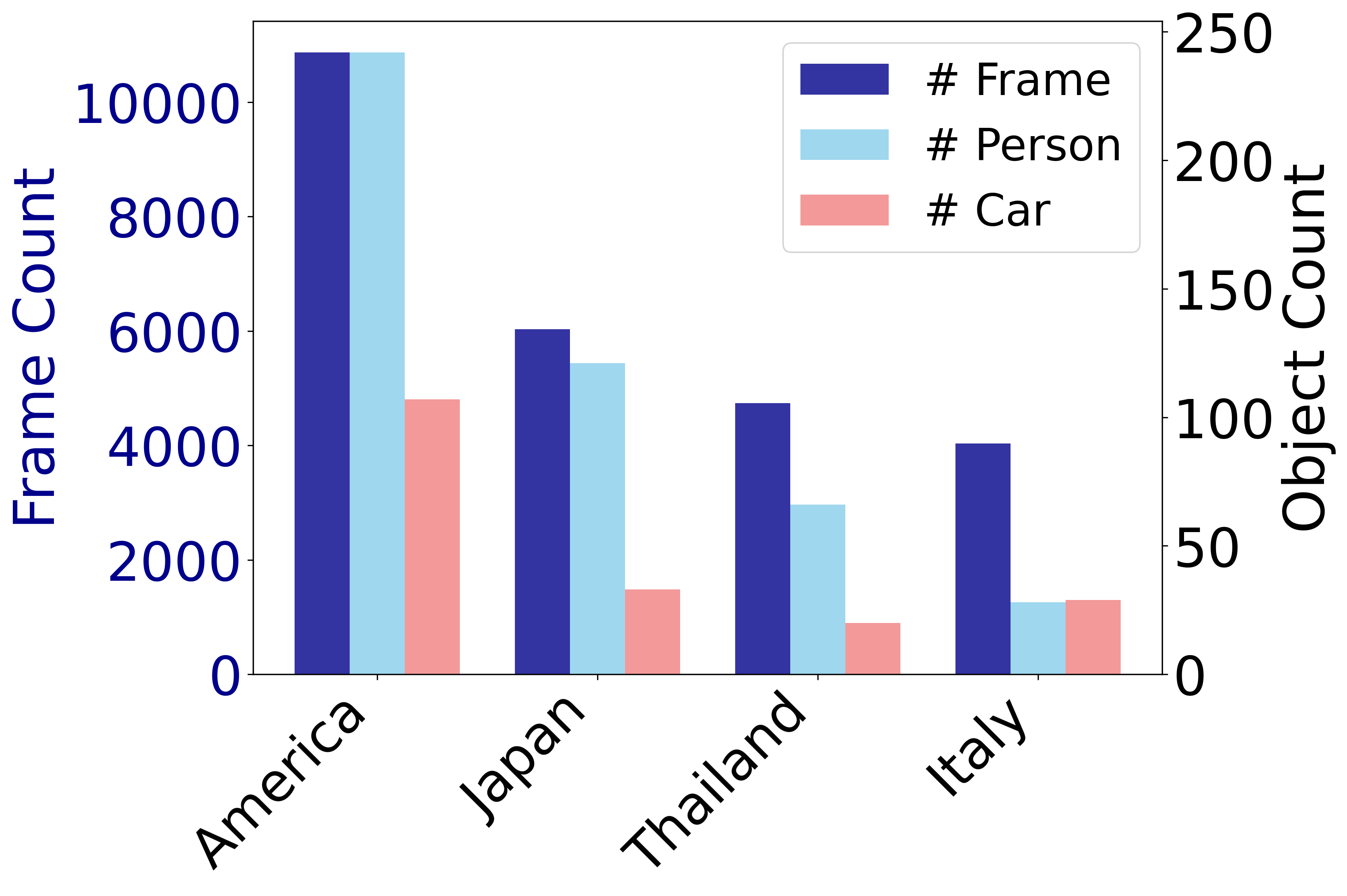}
}
\hfill
\subfloat[Season distribution]{
  \includegraphics[width=0.23\linewidth]{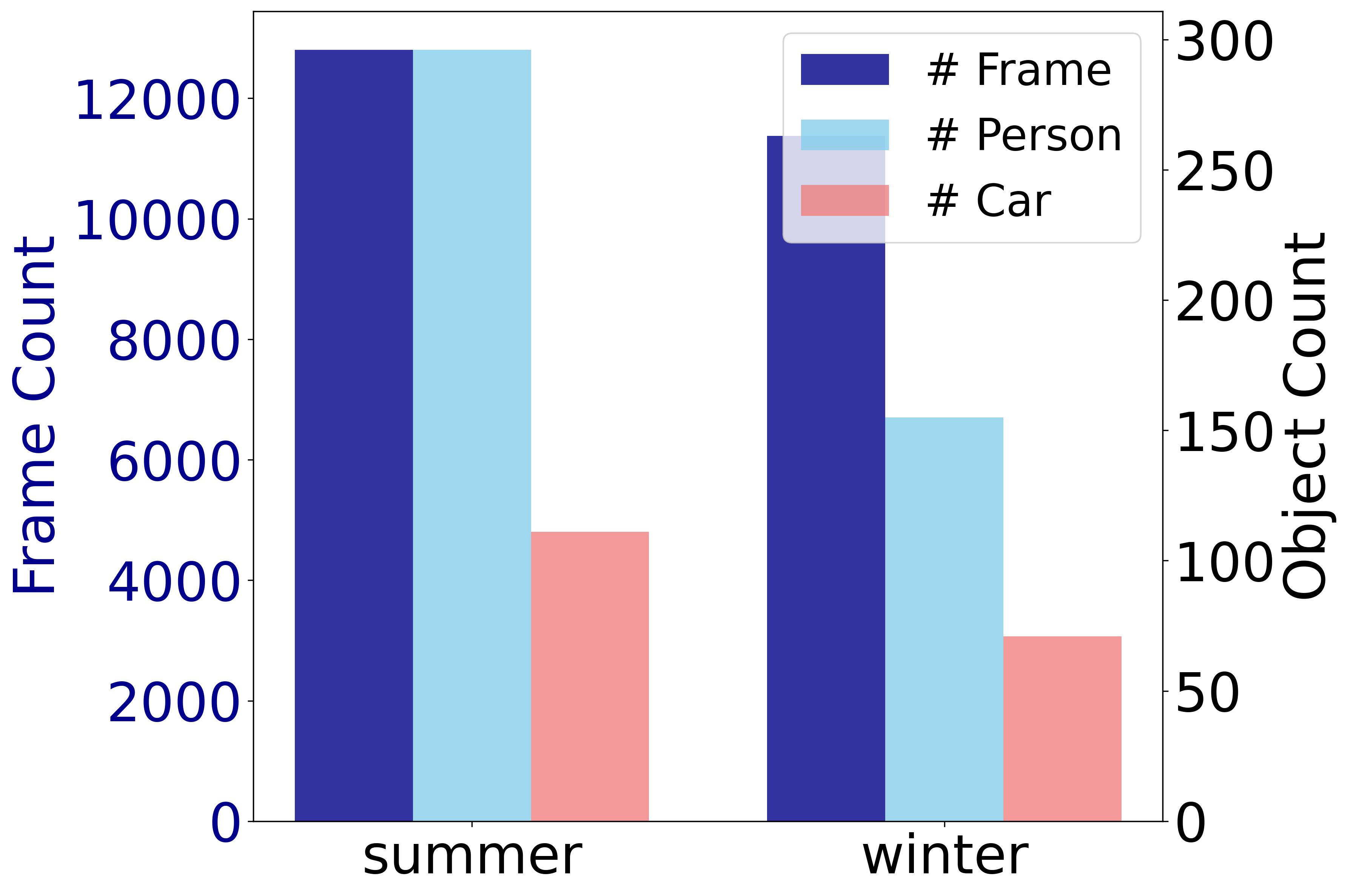}
}
\hfill
\subfloat[Time-of-day distribution]{
  \includegraphics[width=0.23\linewidth]{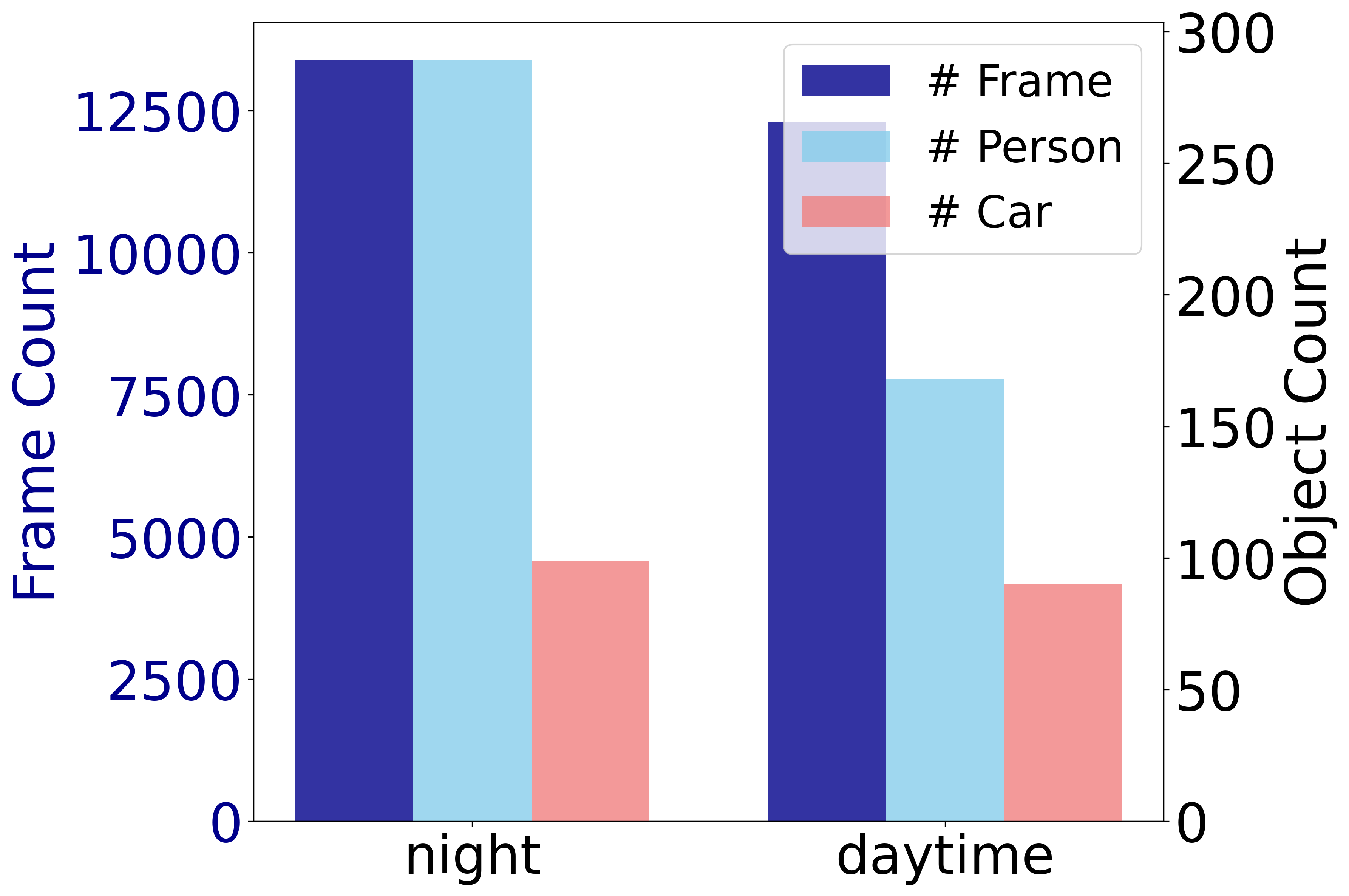}
}
\hfill
\subfloat[Weather distribution]{
  \includegraphics[width=0.23\linewidth]{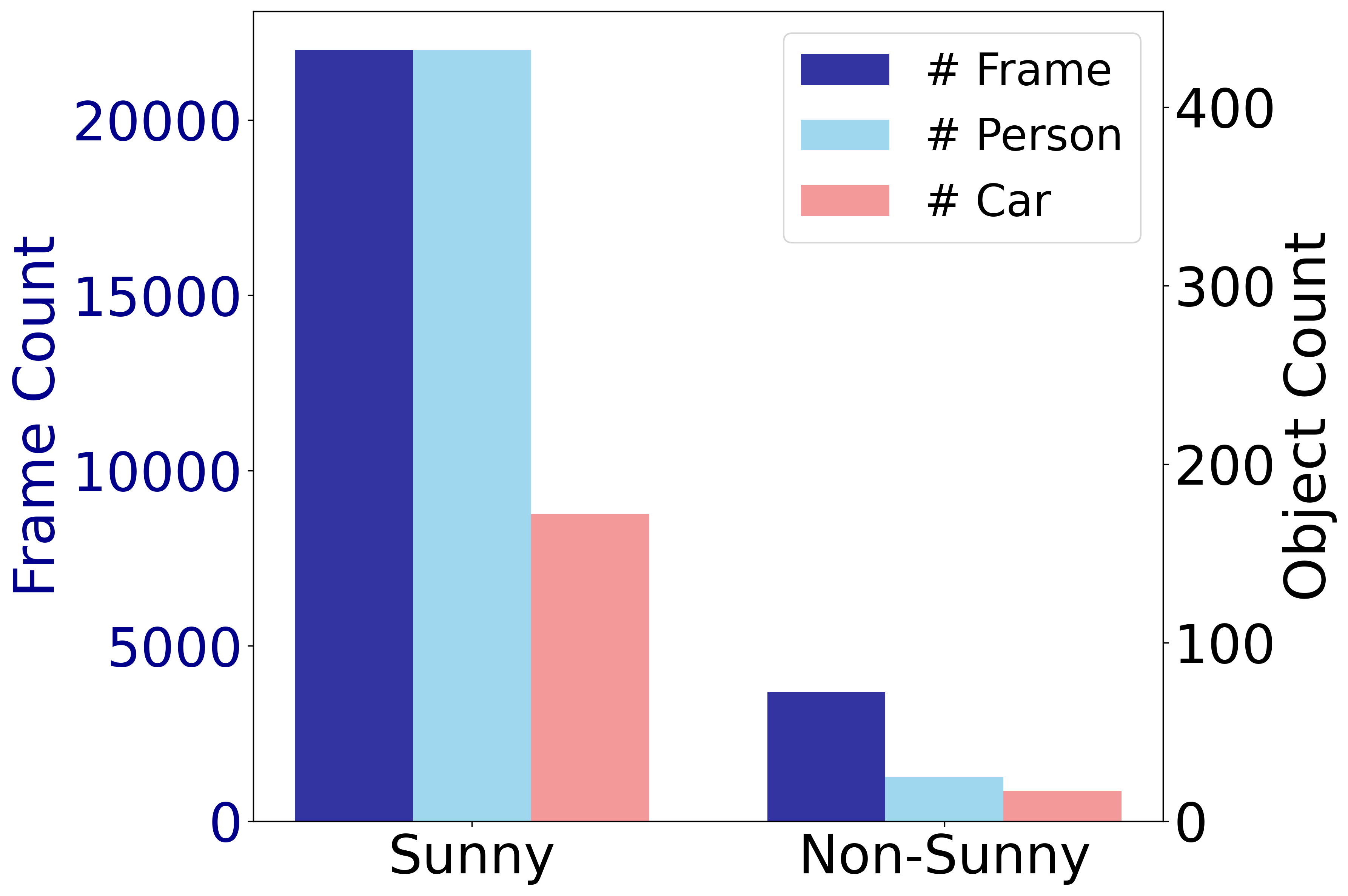}
}
\caption{Distribution of images and annotated targets under different conditions.}
\label{fig:6}
\end{figure*}

After trajectory transformation, we further filter effective segments from long video sequences by retaining only time intervals with dense pedestrian–vehicle interactions and removing redundant segments that contain no interaction or only a single type of traffic participant. As shown in Fig. \ref{fig:4} and Fig. \ref{fig:6} considering the highly imbalanced distribution of environmental conditions such as time of day and weather in raw data, we aim to maintain relatively balanced data distributions across different scenes and environmental conditions during filtering. This strategy helps preserve data diversity and mitigates potential data bias.

\subsection{Data Annotation}

In PINNS, pedestrians and vehicles in each video frame are carefully annotated. The annotation process and content follow the specifications defined in the group standards of the Chinese Association of Automation \cite{TCAA2022002}. Specifically, the dataset comprises over 500 sequences and 50,000 individual trajectory points, fulfilling the scale requirements mandated by the T/CAA standard. Each sequence maintains a minimum duration of 16 frames with a high temporal resolution of 30 Hz. PINNS significantly exceeds the standard’s minimum threshold for trajectory length and frequency (16 frames at 8 Hz), thereby providing more granular motion information for behavior analysis. Furthermore, in strict accordance with the standard's annotation protocols, each object is meticulously labeled with its specific category, precise bounding boxes, and unique tracking IDs to ensure robust cross-frame data association. As illustrated in Fig. \ref{fig:1}, different annotation strategies are adopted for different types of traffic participants:
\begin{itemize}
    \item Pedestrian objects are annotated with bounding boxes in the original image view. The center points of pedestrians on the ground plane are extracted and transformed into the BEV to obtain trajectories in the world plane.
    \item Vehicle objects are annotated with bottom-oriented rotated bounding boxes in the BEV, allowing accurate estimation of vehicle center positions, dimensions, and orientations.
\end{itemize}

The annotation process follows a semi-automatic pipeline. Initial bounding boxes and trajectories are pre-generated using state-of-the-art deep learning-based object detection and multi-object tracking (MOT) methods\cite{bytetrack, 3dvehicle}. These automated annotations are then rigorously reviewed and manually corrected frame-by-frame. To focus on dynamic interactions, stationary objects are excluded as they are treated as environmental background rather than active agents. The final annotation files contain not only world-coordinate trajectories for each object across all frames, but also attributes such as object category, size, and orientation. In addition, for each selected pedestrian–vehicle interaction scene, scene-level information—including location, acquisition season, illumination condition, and weather—is annotated according to the group standards\cite{TCAA2022002}. Beyond the annotated data, we also provide origin view and bird’s-eye-view video clips corresponding to each scene segment, as well as the homography calibration files. Furthermore, scene map annotations will also be provided in future updates of the dataset.

\subsection{Data Statistics}

\begin{figure*}[t]
\centering
\captionsetup[subfloat]{justification=centering, singlelinecheck=false}
\subfloat[Wyoming, America,\\ crossroad]{
  \includegraphics[width=0.23\linewidth]{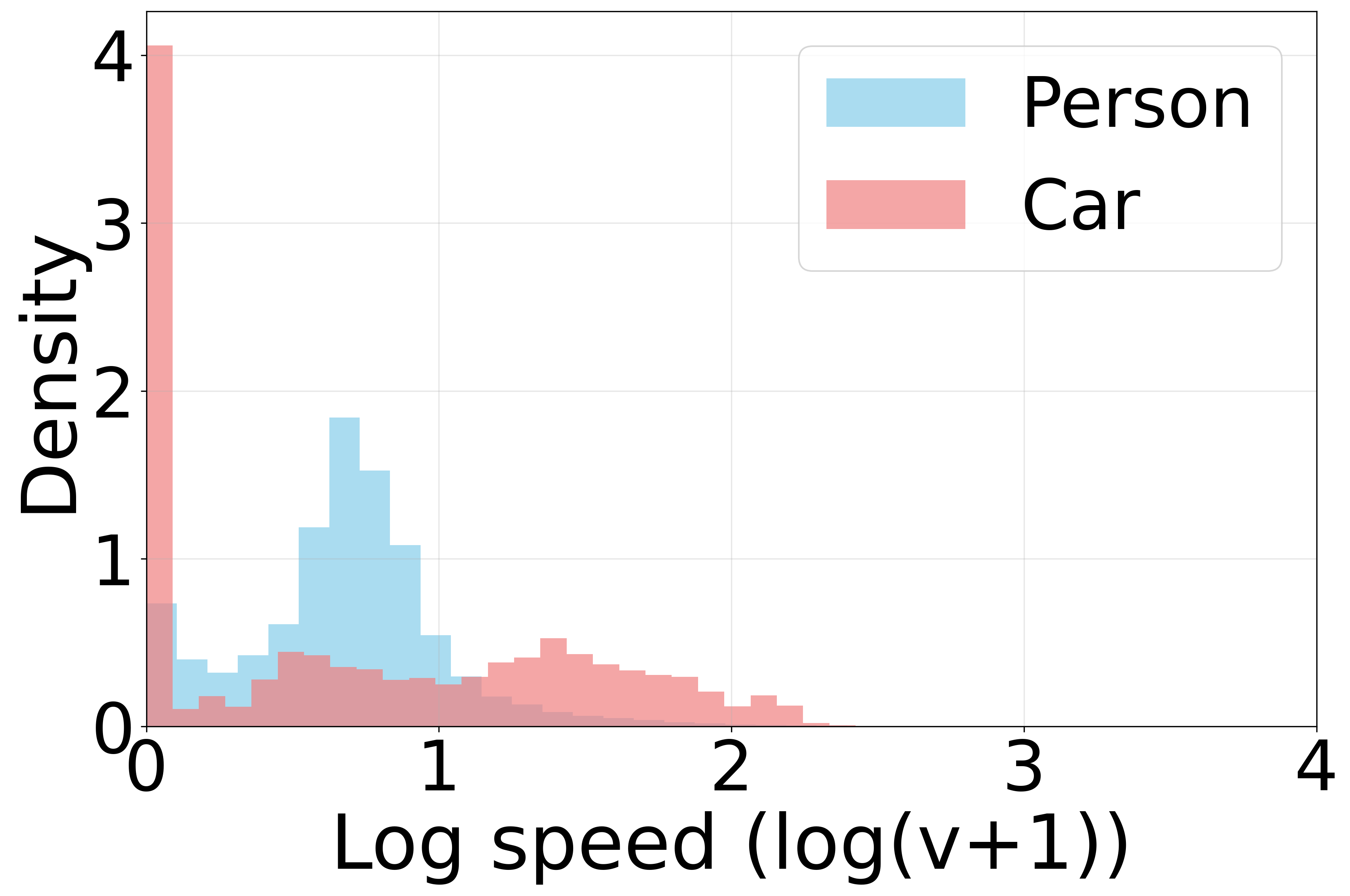}
}
\hfill
\subfloat[Tokyo, Japan,\\ crossroad]{
  \includegraphics[width=0.23\linewidth]{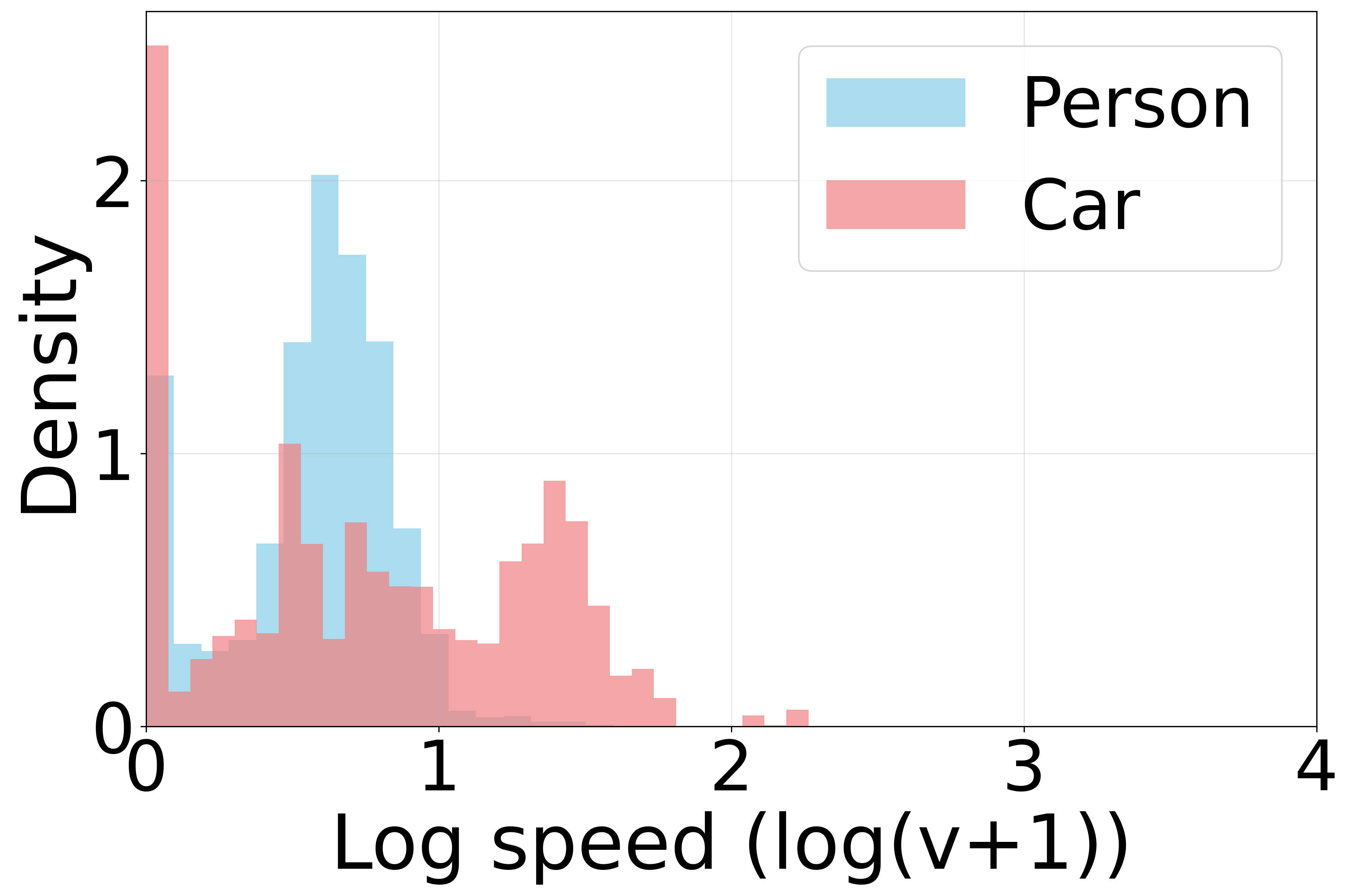}
}
\hfill
\subfloat[Pistoia, Italy,\\ park]{
  \includegraphics[width=0.23\linewidth]{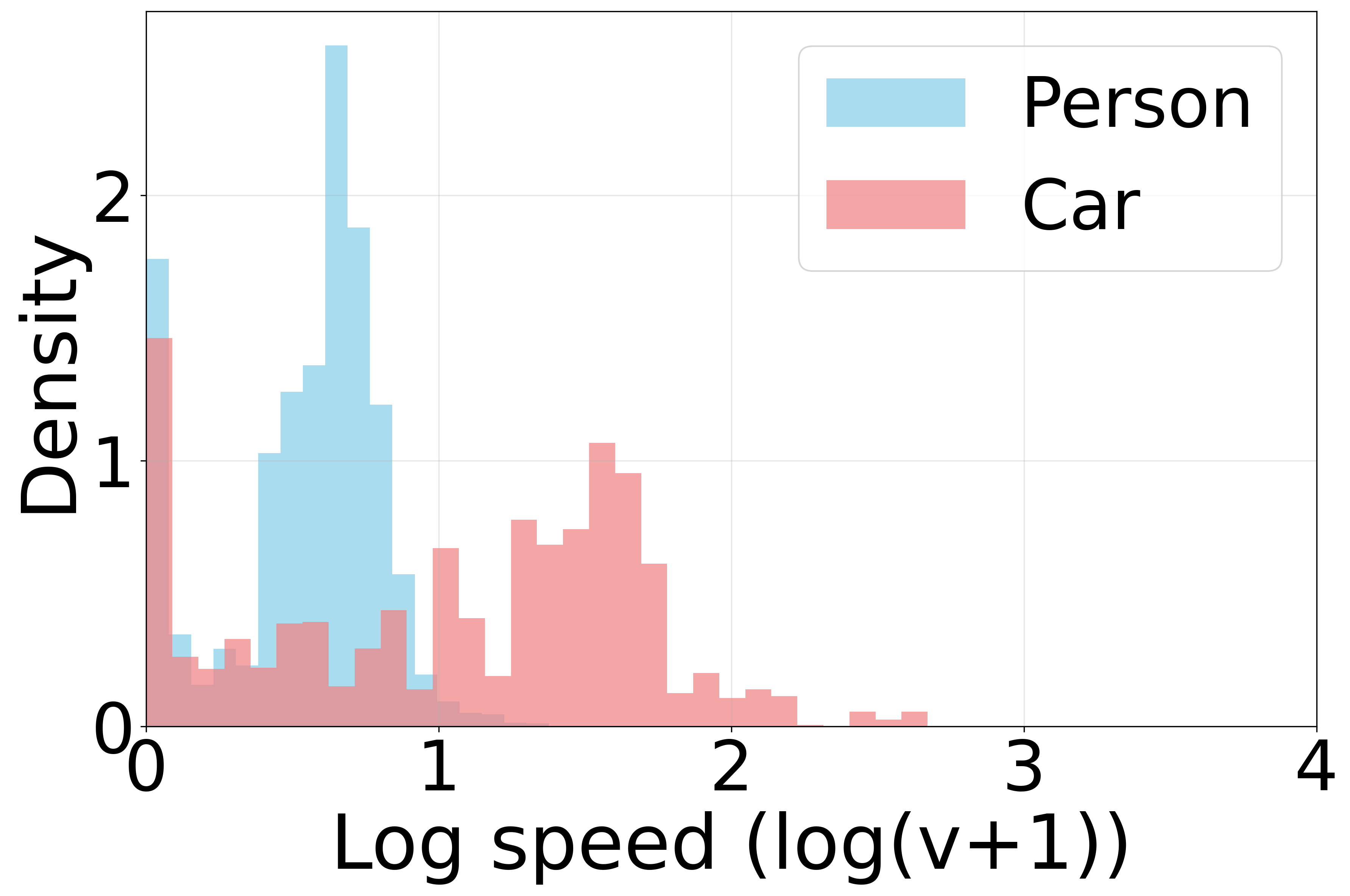}
}
\hfill
\subfloat[Surat Thani, Thailand,\\ street]{
  \includegraphics[width=0.23\linewidth]{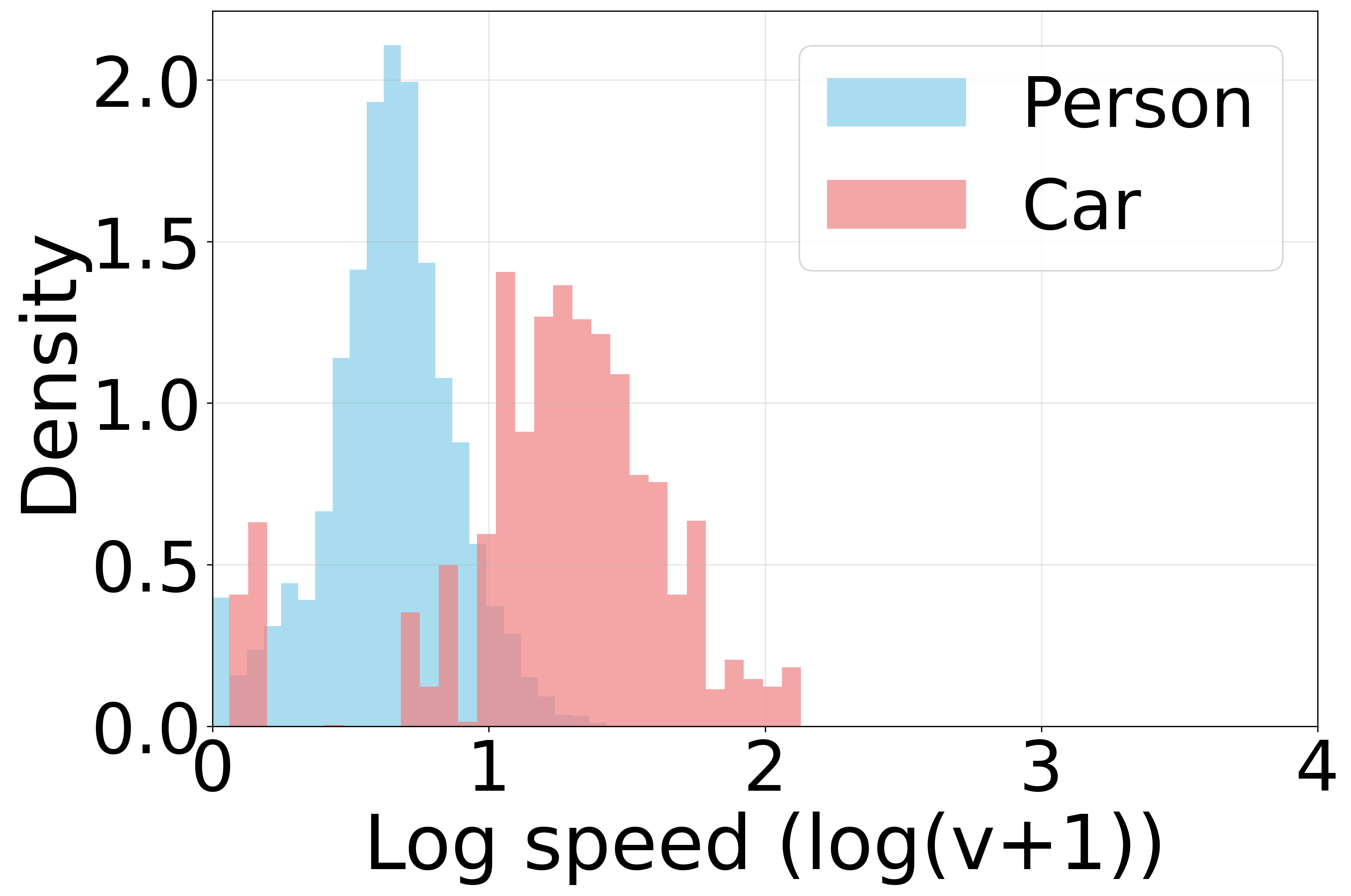}
}
\caption{Speed distributions of pedestrians and vehicles in different scenes and regions.}
\label{fig:7}
\end{figure*}

\begin{figure*}[t]
\centering
\captionsetup[subfloat]{justification=centering, singlelinecheck=false}
\subfloat[Wyoming, America,\\crossroad]{
  \includegraphics[width=0.23\linewidth]{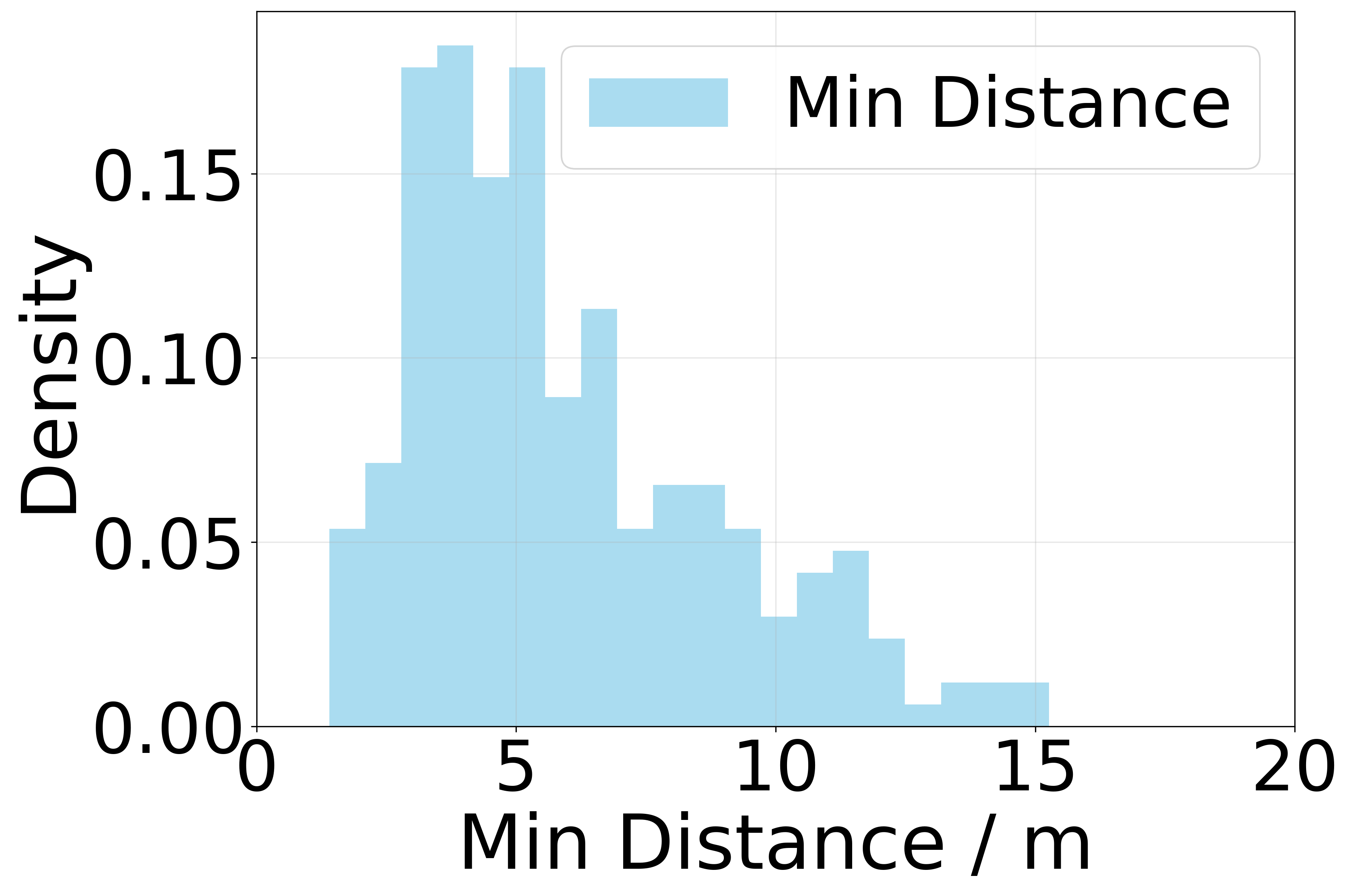}
}
\hfill
\subfloat[Tokyo, Japan,\\crossroad]{
  \includegraphics[width=0.23\linewidth]{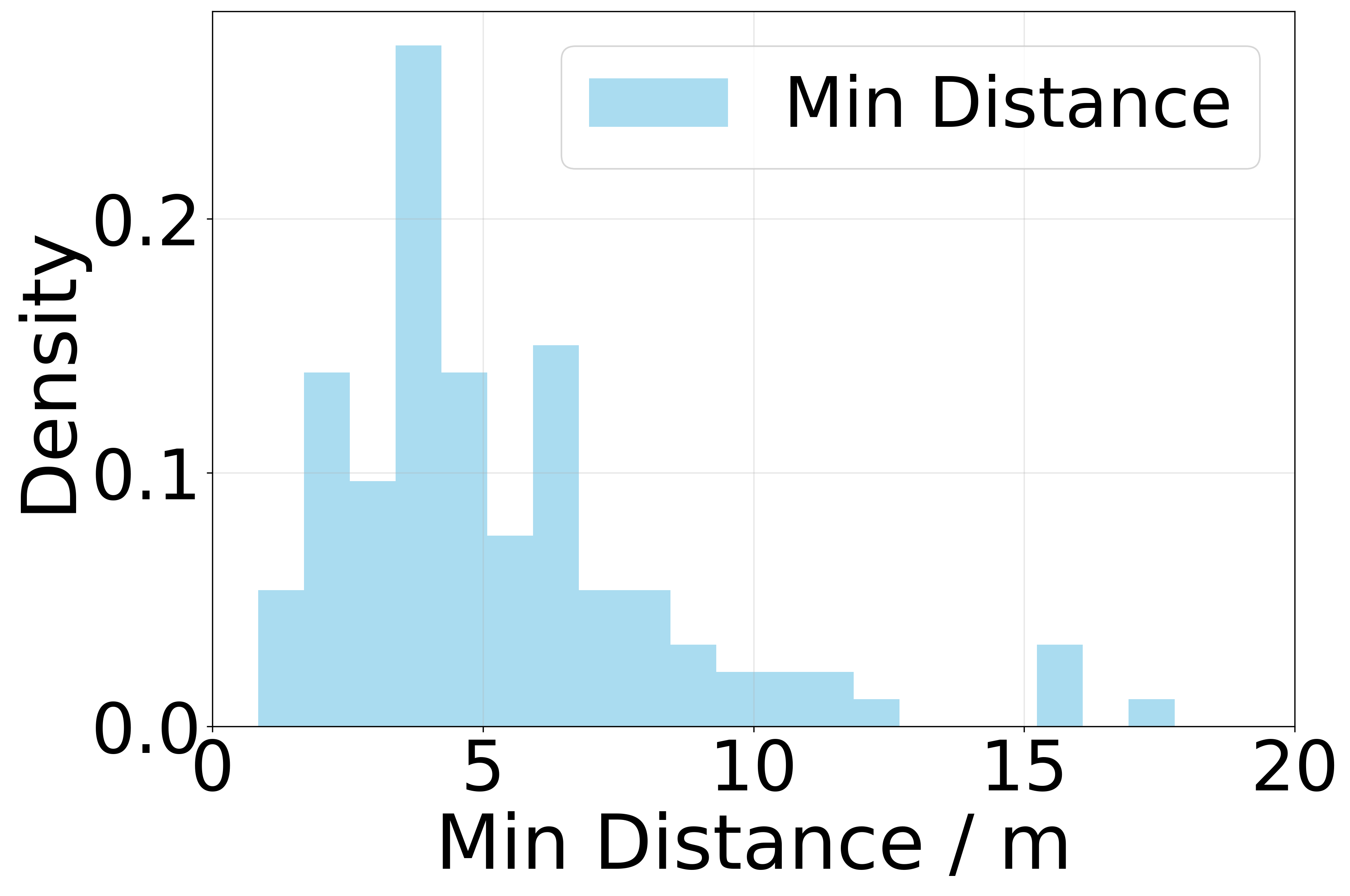}
}
\hfill
\subfloat[Pistoia, Italy,\\park]{
  \includegraphics[width=0.23\linewidth]{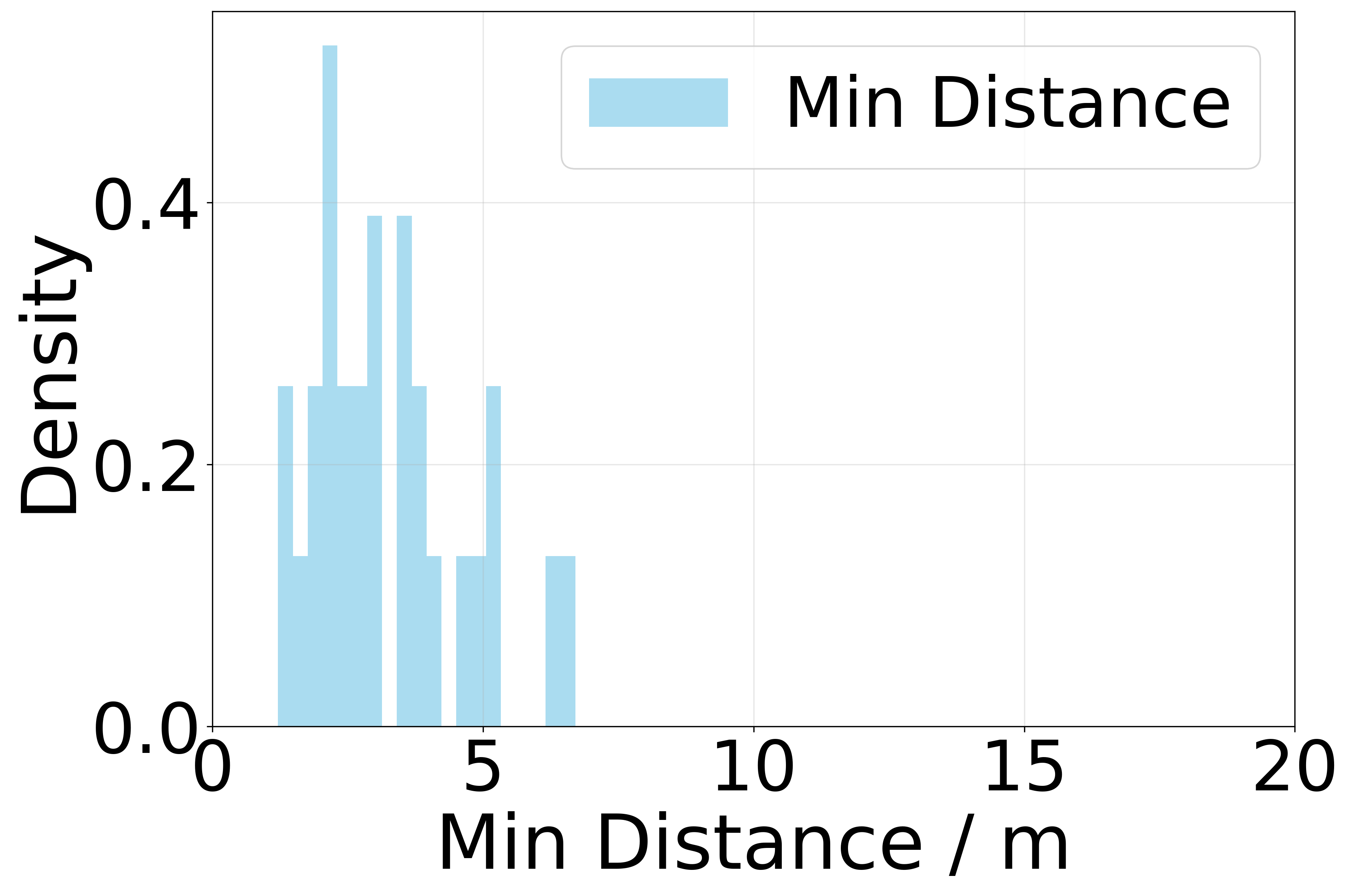}
}
\hfill
\subfloat[Surat Thani, Thailand,\\street]{
  \includegraphics[width=0.23\linewidth]{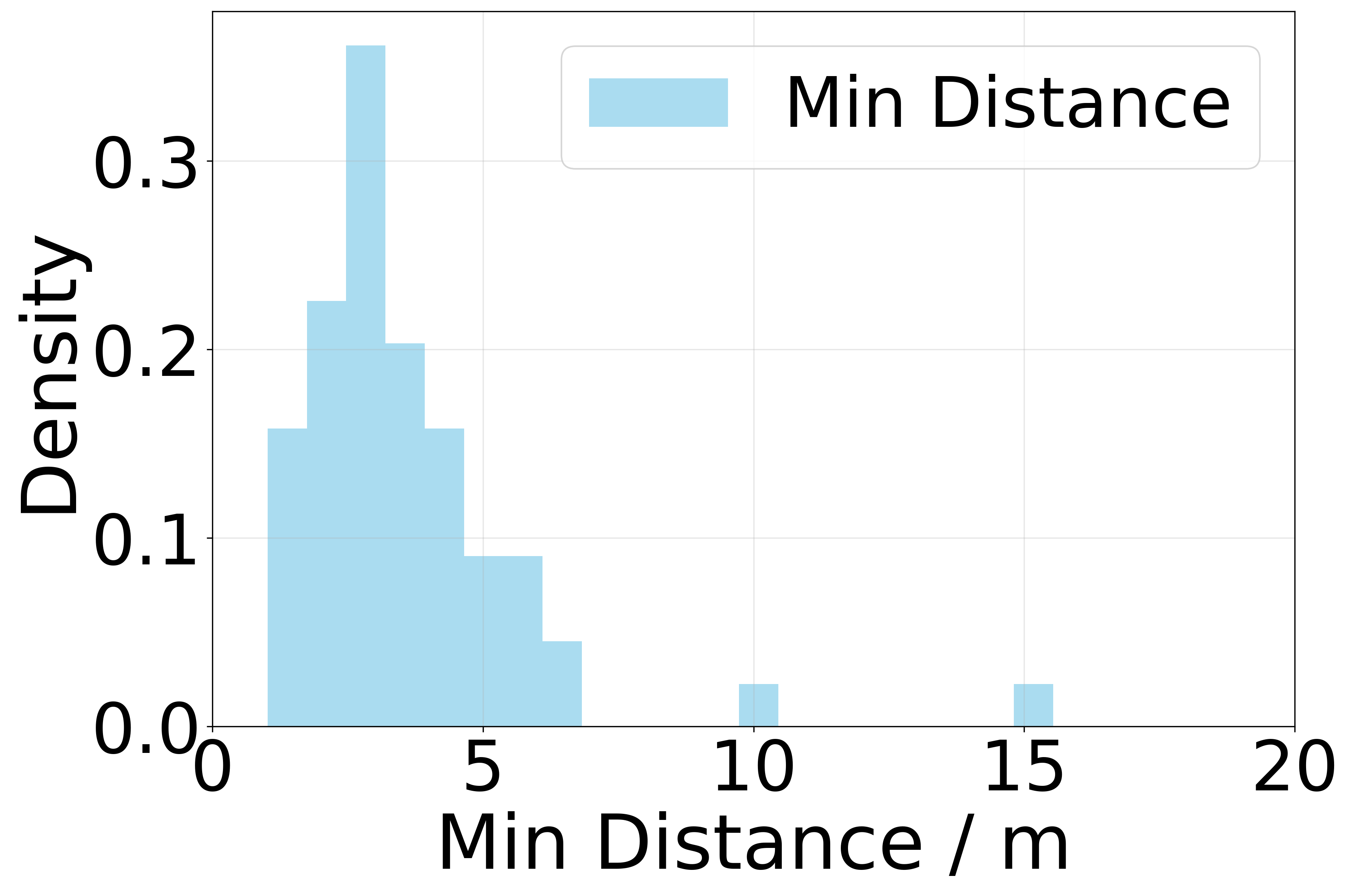}
}
\caption{Distributions of the minimum distances between pedestrians and vehicles in different scenes and locations.}
\label{fig:8}
\end{figure*}

PINNS contains a total of 646 annotated objects, including 457 pedestrians (180344 trajectory points) and 189 vehicles (47555 trajectory points), resulting in 227899 trajectory points. Considering the differences in distribution density, motion speed and scene dwell time between pedestrians and vehicles, the dataset is constructed to maintain comparable orders of magnitude for pedestrian and vehicle annotations.

Further analysis across different scenes, as shown in Fig. \ref{fig:6}, indicates significant differences in the number of observable annotations within the same time duration. These differences are closely related to factors such as location, time, season, and weather, reflecting the diversity and complexity of real-world pedestrian–vehicle interaction behaviors between different scenes. Therefore, instead of enforcing equal annotation counts across scenes, we preserve these differences to avoid distorting the true distribution of interaction scenarios and to maintain dataset diversity.

We further conduct statistical analysis on the collected trajectory data, focusing on the speed distributions of pedestrians and vehicles as well as the distributions of the minimum distance between pedestrians and vehicles. As shown in Fig. \ref{fig:7} and Fig. \ref{fig:8}, which illustrate the speed distributions of pedestrians and vehicles and the nearest-distance distributions under different scenes and regions in logarithmic axis. The results demonstrate significant variations in speed and minimum distance distributions across different scenarios, reflecting distinct behavioral characteristics. In crossroad scenarios, such as America and Japan, the speed distributions for cars exhibit prominent multimodal characteristics. Notably, a sharp peak at near-zero speed indicates frequent waiting behaviors at intersections, while other peaks around $0.5-1.5$ represents the passing phase. This spatial-temporal separation is further evidenced in Fig. \ref{fig:8}(a) and (b), where the minimum interaction distances are relatively large and widely distributed, with peaks occurring between 3m and 8m, suggesting a more structured and regulated interaction pattern.

In contrast, street scenarios like Surat Thani, Thailand present a different dynamic. The speed distributions typically show a more continuous spread, with pedestrian speeds peaking around $0.5$ and the duration of stationary states significantly reduced compared to other structured environments. Concurrently, the minimum distance distributions in these shared-space environments are much more constrained. The high-density occurrence of near-range, non-stationary interactions, combined with the substantial overlap between pedestrian and vehicle speed ranges, intrinsically reflects the profound behavioral coupling and highly complex dynamic features characterizing pedestrian-vehicle interactions in dense urban streets.

Furthermore, even within similar scenario types, regional nuances are evident. Comparing Wyoming, America and Tokyo, Japan, while both involve pedestrian-vehicle interaction in crossroad, the distance distribution in Japan is more concentrated at lower range. These variations are attributable to differences in scene geometry, local traffic regulations, and social norms. Beyond the aforementioned analysis, similar behavioral patterns and distributional disparities are likely to exist across other dimensions, such as varying weather conditions, seasonal transitions, and illumination intensities. Given that these multi-dimensional factors further intertwine to increase the uncertainty of traffic participants' intentions, a more profound quantitative analysis of these aspects is reserved for future research. Such cross-dimensional and multi-scenario data manifestations not only fully demonstrate the diversity of the proposed dataset’s coverage but also underscore the necessity of constructing trajectory datasets that encompass a broad range of environmental conditions to accurately capture complex traffic interaction behaviors.

\section{Benchmark Task and Baseline}
\label{sec:4}
In this section, we introduce the benchmark task targeted by PINNS, namely trajectory prediction in pedestrian–vehicle interaction scenarios. We describe the task formulation and evaluation metrics in detail, and select representative methods as baseline to conduct experimental validation and result analysis on PINNS.

\subsection{Pedestrian–Vehicle Trajectory Prediction}

The total number of pedestrians and vehicles appearing in a scene is denoted as $N$. At each time step $t$, the position of a pedestrian or vehicle in the world coordinate system is represented as $(x^i_t, y^i_t), i\in\{1,\dots,N\}$. Given the historical trajectory sequences of all pedestrians and vehicles within the observation interval $t = 1, \dots, T_{obs}$, namely $(x^i_t, y^i_t), i\in\{1,\dots,N\},t\in\{1, \dots, T_{obs}\}$, the goal of the pedestrian–vehicle trajectory prediction task is to predict the corresponding future trajectories during the prediction interval $t=T_{obs}+1, \dots, T_{pred}$, this is $(x^i_t, y^i_t), i\in\{1,\dots,N\},t\in\{T_{obs}+1, \dots, T_{pred}\}$. In addition to the position sequences, auxiliary information such as motion direction, velocity, and scene-level attributes can be optionally provided as additional inputs to improve prediction accuracy.

\subsection{Evaluation Metrics}
We evaluate the prediction performance using two metrics widely adopted in the trajectory prediction literature: Average Displacement Error (ADE) and Final Displacement Error (FDE).

The ADE measures the average Euclidean distance between the predicted trajectory and the ground-truth over the entire prediction horizon, defined as:
\begin{equation}
\label{eq:ade}
\mathrm{ADE} = \frac{1}{T_{\mathrm{pred}}}\sum_{t=T_{\mathrm{obs}}+1}^{T_{\mathrm{obs}}+T_{\mathrm{pred}}}\left|\left| \hat{\mathbf{p}}_t - \mathbf{p}_t \right|\right|_2,
\end{equation}
where $\mathbf{p}_t = (x_t, y_t)$ denotes the ground-truth coordinates at time step $t$, and $\hat{\mathbf{p}}_t = (\hat{x}_t, \hat{y}_t)$ represents the corresponding predicted position.

The FDE calculates the Euclidean distance between the predicted final position and the ground-truth final destination at the end of the prediction period, formulated as:
\begin{equation}
\label{eq:fde}
\mathrm{FDE} = \left|\left| \hat{\mathbf{p}}_{T{\mathrm{obs}}+T_{\mathrm{pred}}} -\mathbf{p}_{T{\mathrm{obs}}+T_{\mathrm{pred}}} \right|\right|_2.
\end{equation}

In Eq. \ref{eq:ade} and Eq. \ref{eq:fde}, $T_{\mathrm{obs}}$ and $T_{\mathrm{pred}}$ denote the number of observed and predicted time steps, respectively.

\subsection{Baseline and Results}

\begin{table*}[t]
\centering
\caption{Performance of baseline model on different datasets.}
\label{tab:ade_fde_comparison}

\small
\setlength{\tabcolsep}{10pt}
\renewcommand{\arraystretch}{1.2}

\begin{tabular}{l p{1.3cm} p{1.3cm} p{1.3cm} p{1.3cm} p{1.3cm} p{1.8cm} p{1.8cm}}
\hline
\textbf{Metric} & \textbf{ETH (Ped)\cite{eth}} & \textbf{Hotel (Ped)\cite{eth}} & \textbf{Univ (Ped)\cite{ucy}} & \textbf{Zara1 (Ped)\cite{ucy}} & \textbf{Zara2 (Ped)\cite{ucy}} & \textbf{nuScenes (Ped/Veh)\cite{nuscenes}} & \textbf{PINNS (Ped/Veh)} \\
\hline
ADE$\downarrow$ & 0.71 & 0.22 & 0.44 & 0.30 & 0.23 & -- & 0.96 / 1.90 \\
FDE$\downarrow$ & 1.66 & 0.46 & 1.17 & 0.79 & 0.59 & 0.60 / 2.20 & 1.93 / 3.79 \\
\hline
\end{tabular}
\par
\smallskip
\begin{minipage}{0.916\textwidth} %
    \footnotesize Ped represents pedestrian, Veh represents vehicle. ADE of nuScenes is not available in previous works.
\end{minipage}
\end{table*}

We select Trajectron++\cite{trajectron} as the baseline model to evaluate the proposed dataset. Trajectron++ is a representative heterogeneous multi-agent trajectory prediction method. It adopts LSTM-based sequence modeling and leverages a dynamic graph structure together with a probabilistic generative framework to jointly model interactions among different types of traffic participants.

To ensure comparability with existing work, we adopt the same model architecture, input and output settings as previous studies. Therefore, before evaluation, we downsample our dataset to 2.5 Hz to ensure consistency in the time length. We use an observation length of 8 timesteps (3.2 s) and a prediction length of 12 timesteps (4.8 s), following the standard experimental protocol used in prior work. According to the requirements of the group standard released by the Chinese Association of Automation \cite{TCAA2022002}, we conduct experiments on PINNS using three-fold cross-validation, and use the average performance over all folds as the final evaluation metric. Meanwhile, we compare the performance of the evaluated method on PINNS with the results of previously published datasets. 

The quantitative experimental results, summarized in Table \ref{tab:ade_fde_comparison}, indicate that achieving precise trajectory prediction on PINNS presents a significantly greater challenge than on conventional benchmarks. Specifically, the ADE and FDE for pedestrians in PINNS reach $0.96$ / $1.93$, which are markedly higher than those observed in other widely used datasets. We posit that this elevated prediction error stems not only from the structural design of existing algorithms, which tend to favor structured scenarios and simplified interactions, but more fundamentally from the complexity and uncertainty embedded within the data itself. This performance disparity is further exacerbated in vehicle trajectories, due to their higher velocities and more volatile motion profiles, where the FDE reaches 3.79, nearly double the error recorded for pedestrians. Consistent with the distributions illustrated in Fig. \ref{fig:7} and Fig. \ref{fig:8}, PINNS captures a substantially broader spectrum of motion dynamics and interaction modalities. Unlike the ETH/UCY datasets\cite{eth, ucy}, which are primarily characterized by homogeneous pedestrian interactions in open spaces, PINNS encompasses a vast array of complex, close-range encounters between pedestrians and vehicles. In such mixed-traffic environments, the intense spatiotemporal coupling between heterogeneous agents significantly enhances the multimodal nature of the prediction task. In summary, these results demonstrate that real-world, heterogeneous, and close-range interaction data possesses a higher benchmark difficulty. This validates the necessity and primary objective of PINNS: to provide a more representative foundation for solving complex pedestrian-vehicle interaction trajectory prediction problems in unstructured scenarios.

\section{Conclusion}
\label{sec:5}
In this paper, we address the lack of trajectory prediction data for complex pedestrian–vehicle interaction scenarios in semi-structured and unstructured road environments. We propose a dataset construction method based on video data from uncalibrated surveillance cameras and build a pedestrian–vehicle interaction trajectory prediction dataset targeting semi-structured and unstructured scenarios. The dataset covers multiple regions, diverse scene types, and varied environmental conditions, and the proposed framework demonstrates strong extensibility. We choose and evaluate representative heterogeneous agent trajectory prediction baselines on the dataset, showing that existing methods suffer a significant performance degradation compared with their results on conventional datasets. In particular, notable deficiencies are observed in vehicle trajectory prediction and heterogeneous interaction modeling, indicating that current approaches still have limited generalization capability in complex traffic environments. These results highlight the importance of constructing high-quality trajectory prediction datasets that faithfully capture complex pedestrian–vehicle interactions, in order to advance the development and evaluation of trajectory prediction methods in mixed traffic scenarios. We hope that the proposed dataset will benefit future research on interaction modeling and trajectory prediction in mixed environments, enhance traffic safety of autonomous driving, and promote the deployment of autonomous driving systems in complex real-world settings.

\bibliographystyle{IEEEtran}
\bibliography{references}

\section{Biography}
\begin{IEEEbiography}[{\includegraphics[width=1in,height=1.25in,clip,keepaspectratio]{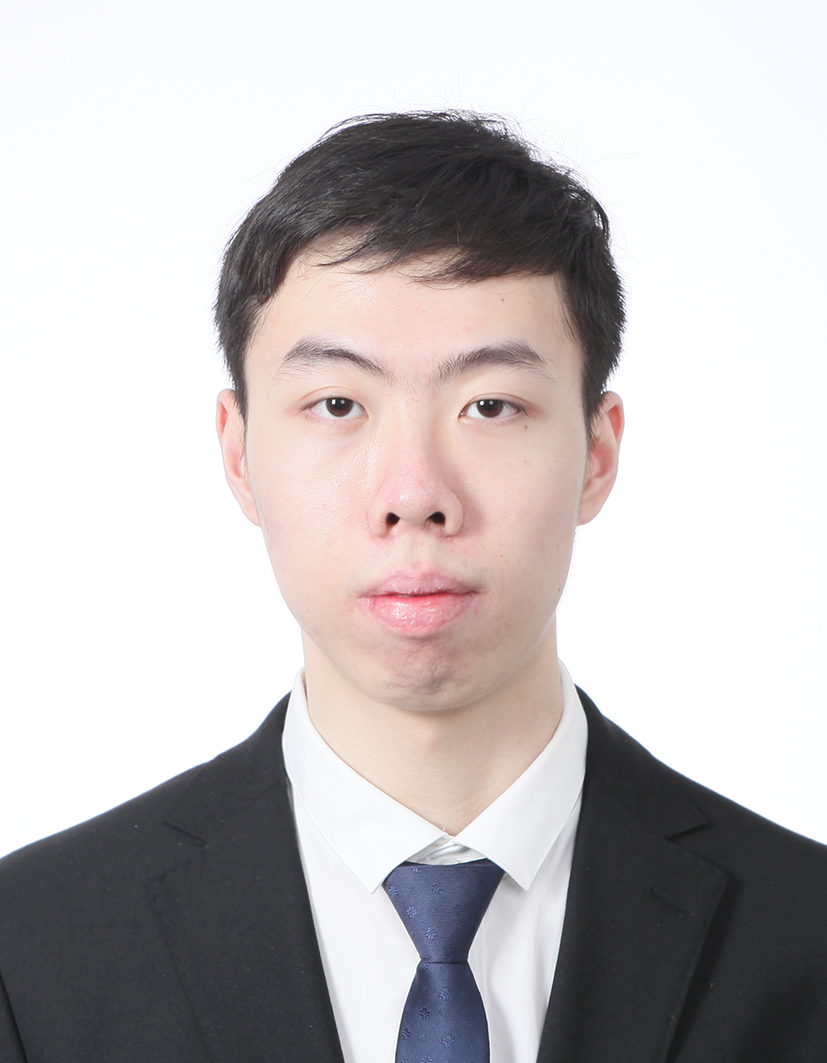}}]{Haoyang Peng}
received the B.S. degree in Artificial Intelligence from Shanghai Jiao Tong University, Shanghai, China, in 2025, where he is currently pursuing the M.S. degree in Electronic Information Engineering. His main research interests include deep learning and autonomous driving.
\end{IEEEbiography}

\vfill

\begin{IEEEbiography}[{\includegraphics[width=1in,height=1.25in,clip,keepaspectratio]{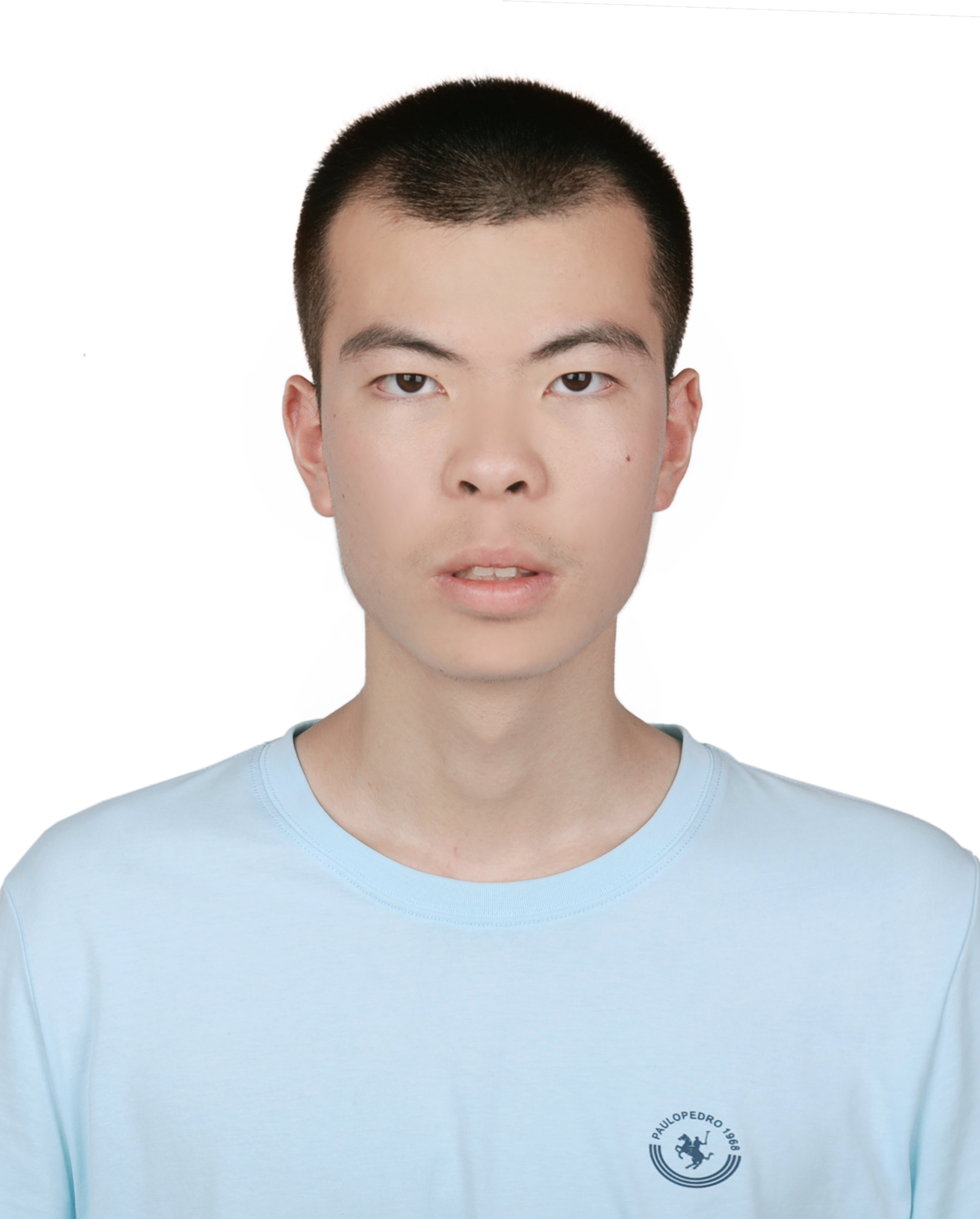}}]{Qian Hu}
is a sophomore of Global Institute of Future Technology (GIFT), Shanghai Jiao Tong University, Shanghai, China, where he is currently pursuing the B.S. degree in Sustainable Energy Technology. His main research interests include deep learning and autonomous driving.
\end{IEEEbiography}

\vfill
\begin{IEEEbiography}
[{\includegraphics[width=1in,height=1.25in,clip,keepaspectratio]{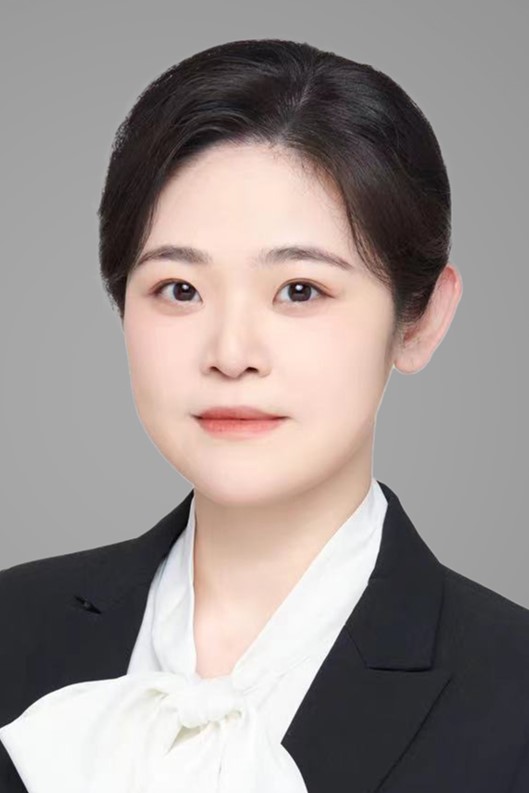}}]{Songan Zhang}
received the B.S. and M.S. degrees in automotive engineering from Tsinghua University, Beijing, China, in 2013 and 2016, respectively. In 2021, she earned the Ph.D. degree in mechanical engineering from the University of Michigan, Ann Arbor, MI, USA. Upon graduation, she joined Ford Motor Company as a Research Scientist where she made contributions to pioneering innovations in smart manufacturing and advanced driver assist systems. She is currently an Assistant Professor with the Global Institute of Future Technology, Shanghai Jiao Tong University, Shanghai, China. Her research interests include accelerated and safety evaluation of autonomous vehicles; verification methods for autonomous driving systems; model-based, trustworthy, and data-efficient reinforcement learning and meta-learning; and the application of foundation models for decision-making in autonomous vehicles and intelligent transportation systems.
\end{IEEEbiography}

\vfill

\begin{IEEEbiography}
[{\includegraphics[width=1in,height=1.25in,clip,keepaspectratio]{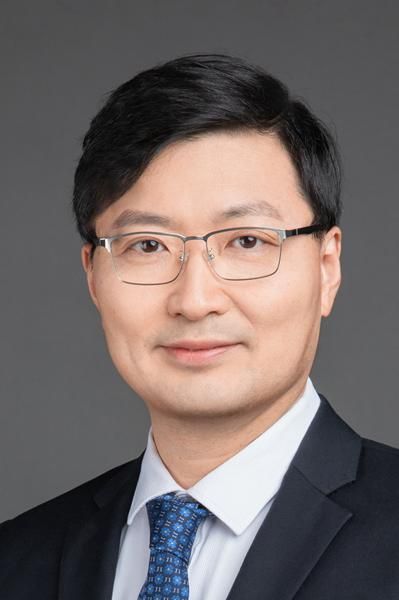}}]{Ming Yang}
received the Master and Ph.D. degrees from Tsinghua University, Beijing, China, in 1999 and 2003, respectively. He is currently the Full Tenure Professor at Shanghai Jiao Tong University, the deputy director of the Innovation Center of Intelligent Connected Vehicles. He has been working in the field of intelligent vehicles for more than 20 years. He participated in several related research projects, such as the THMR-V project (first intelligent vehicle in China), European CyberCars and CyberMove projects, CyberC3 project, CyberCars-2 project, ITER transfer cask project, AGV, etc.
\end{IEEEbiography}

\end{document}